\RequirePackage{fix-cm}
\documentclass[smallextended, final]{svjour3}       % onecolumn (second format)
\journalname{Data Mining and Knowledge Discovery}
\smartqed  % flush right qed marks, e.g. at end of proof
\usepackage{graphicx}
\usepackage[square,sort,comma,numbers]{natbib}

\usepackage{comment}
\usepackage{hyperref}
\usepackage{booktabs}
\begin{document}

\title{Fast and Robust Video-Based Exercise Classification via Body Pose Tracking and Scalable Multivariate Time Series Classifiers %\thanks{Grants or other notes
}

\titlerunning{Fast and Robust Video-Based Exercise Classification}        % if too long for running head

\author{Ashish Singh    \and
        Antonio Bevilacqua \and 
        Thach Le Nguyen \and 
        Feiyan Hu \and
        Kevin McGuinness \and
        Martin O'Reilly \and 
        Darragh Whelan \and 
        Brian Caulfield \and 
        Georgiana Ifrim%etc.
}


\institute{Ashish Singh, Antonio Bevilacqua, Thach Le Nguyen and Georgiana Ifrim are with the Insight Centre for Data Analytics, School of Computer Science, University College Dublin, Dublin, Ireland;
        Kevin McGuinness and Feiyan Hu are with  Insight Centre for Data Analytics, School of Electronic Engineering, Dublin City University, Dublin, Ireland;
        Brian Caulfield is with the Insight Centre for Data Analytics, School of Public Health, Physiotherapy and Sports Science, University College Dublin, Dublin, Ireland \\
        \{ashish.singh,antonio.bevilacqua,thach.lenguyen,kevin.mcguinness,b.caulfield,feiyan.hu, georgiana.ifrim\}@insight-centre.org             \\
    Martin O'Reilly, Darragh Whelan are with Output Sports Limited, NovaUCD, Dublin, Ireland; 
    \{darragh, martin\}@outputsports.com 
}
\date{Received: date / Accepted: date}

\maketitle

\begin{abstract}
    %-------------------------------------------------------------------------
	%The abstract should briefly summarize the contents of the paper in	about 250-300 words.
	%-------------------------------------------------------------------------
	%1. What is the problem addressed?
    %2. Why is it important?
    %3. Why is it challenging?
    %4. What are the existing solutions and their limitations? 
    %5. What does this paper propose? 
    %6. What are the results/lessons learned so far? 
	%1. What is the problem addressed?

%Requirements: (Sect: abstract + intro; rel work)
%Task: Strength and Conditioning exercise classification: why important, why challenging, gaps (we do not focus on classifying dance motion or other sports)
%Video data capture
%Light computation for training/test
%High accuracy
%Robustness to noise

% Background
Recent technological advancements have spurred the usage of machine learning based applications in sports science and healthcare. Using wearable sensors and video cameras to analyze and improve the performance of athletes, has become widely popular.
Physiotherapists, sports coaches and athletes actively look to incorporate the latest technologies in order to further improve performance and avoid injuries. While wearable sensors are very popular, their use is hindered by constraints on battery power and sensor calibration, especially for use cases which require multiple sensors to be placed on the body. 
Hence, there is renewed interest in video-based data capture and analysis for sports science. 
In this paper, we present the application of classifying strength and conditioning exercises using video. We focus on the popular Military Press exercise, where the execution is captured with a video-camera using a mobile device, such as a mobile phone, and the goal is to classify the execution into different types. 
Since video recordings need a lot of storage and computation, this use case requires data reduction, while preserving the classification accuracy and enabling fast  prediction.
To this end, we propose an approach named BodyMTS to turn video into time series by employing body pose tracking, followed by training and prediction using multivariate time series classifiers. We analyze the accuracy and robustness of BodyMTS and show that it is robust to different types of noise caused by either video quality or pose estimation factors. We compare BodyMTS to state-of-the-art deep learning methods which classify human activity directly from videos and show that BodyMTS achieves similar accuracy, but with reduced running time and model engineering effort.
Finally, we discuss some of the practical aspects of employing BodyMTS in this application in terms of accuracy and robustness under reduced data quality and size. We show that BodyMTS achieves an average accuracy of 87\%, which is significantly higher than the accuracy of human domain experts.
\end{abstract}

% Redo the abstract to reflect the answers to DAMI application paper requirements

%%%%%%%%%%%%%%%% INTRO %%%%%%%%%%%%%%%%%%%%%%

\section{Introduction}
\label{sec:intro}
	
	%1. What is the problem addressed?
    %2. Why is it important?
    %3. Why is it challenging?
    %4. What are the existing solutions and their limitations? 
    %5. What does this paper propose? 
    %6. What are the results/lessons learned so far? 
    
    % what do you want to have achieved by the end of your project? (ideal outcome)
    % what is the main technical novelty in your research?
    % what was the biggest challenge in your research? how did you address it?
    % what techniques did you use? what are their strengths and limitations?
    % what is the key contribution of your project? how does it advance the state-of-the-art in a specific area?
    % what are the top conferences/journals in your area?
    % what are the main groups working in your area?
    
    % Overall
    Recent years have seen a tremendous growth of the use of machine learning for sports science and healthcare applications. This is mainly due to the increased usage of wearable sensors and video-based tracking devices ~\citep{ahmadi2014automatic, Martin2015, Martin2017, Martin2018, Fawaz2019, Kwon2020, Potion2018} to capture data that is utilized for rehabilitation or to assess the performance of athletes \citep{Richter2021}. 
    
    Human exercise performance classification is a sub-field of human activity recognition (HAR) where the goal is to classify the execution of an exercise into predetermined classes. 
    %Henceforth, we use the terms ``classification of physical exercises" and ``human exercise performance classification" interchangeably. 
    Most research in this field has focused on utilizing inertial sensors for data capture \citep{ahmadi2014automatic, Martin2015, Martin2017, Martin2018, Fawaz2019}, which commonly involves extracting domain-specific or predefined statistical features from sensor data and applying supervised machine learning methods. However, using sensors to collect human activity data has some notable limitations: sensor-based data collection is  error-prone and time-consuming as sensors require careful positioning on the body, as well as calibration for the specific task \citep{Darragh2016, Kwon2020}.
    
    This work focuses on classifying physical exercise execution by using video data capture. Video data helps to alleviate some of the above problems, as videos can be easily captured through available smartphones and data capture does not require multiple specialized sensor devices to be worn on the body, thus avoiding issues such as discomfort and impending the ease of movement \citep{Kwon2020}. In this paper, we work with video recordings of participants executing the Military Press (MP) exercise. MP is an important exercise in strength and conditioning, injury risk screening, and rehabilitation \citep{Darragh2016}. The main objective is to classify exercise performance in terms of differentiating between correct and different aberrant executions of the exercise. Incorrect execution may lead to muscoskeletal injuries and impede performance \citep{Baechle2008}, therefore, automated and accurate feedback on execution is important to avoid injuries and maximize the performance of the user. While this is an important exercise, it is also a difficult one to classify, with human inter-rater agreement at about 60\% \citep{whelan2019determining}.
    
\begin{figure*}[ht!]
    \centering
        \includegraphics[width=1.0\textwidth]{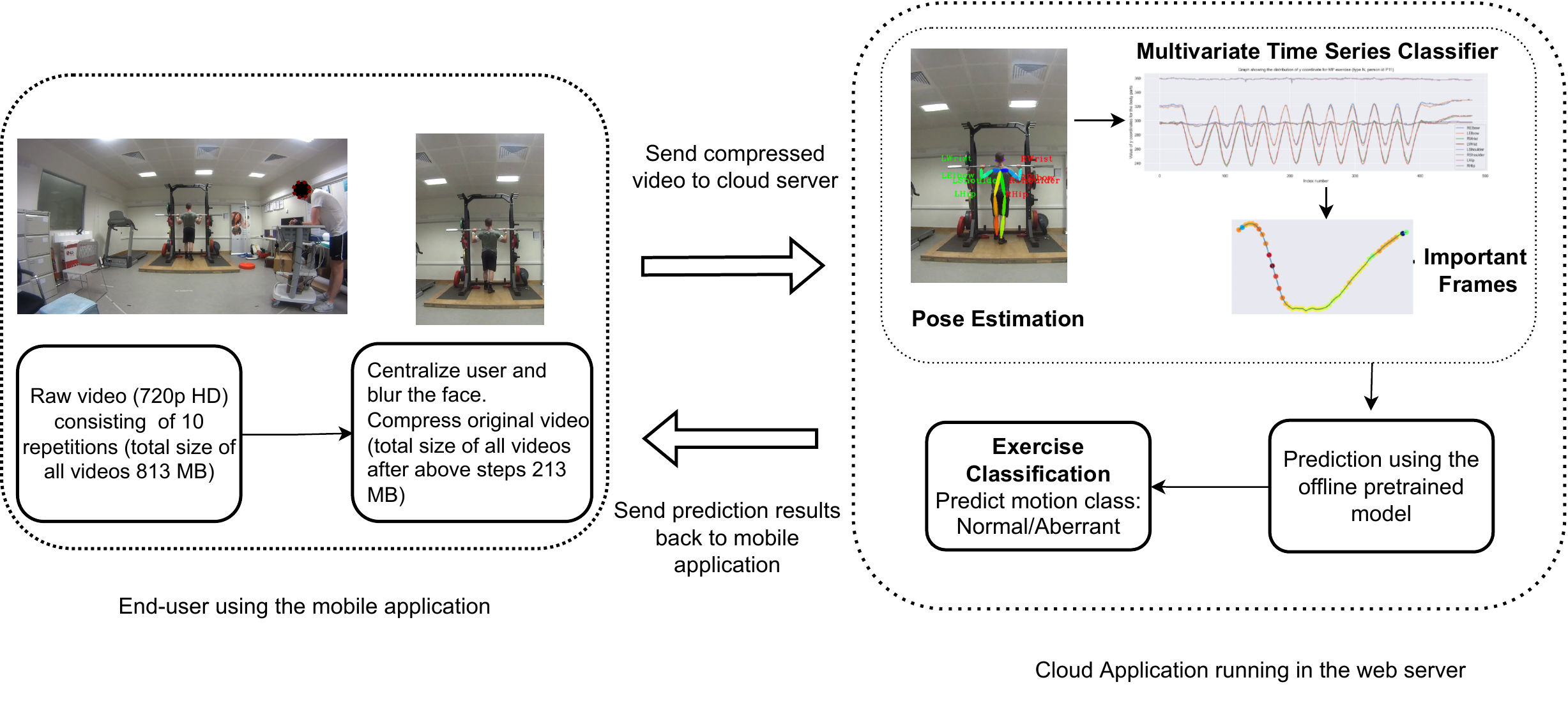}
        \caption{Overview of the BodyMTS approach for the Military Press strength and conditioning exercise classification. Left-to-right flow: raw video, extracting and tracking body parts using human pose estimation, preparing the resulting data for multivariate time series classification and interpretation.}
        \label{fig:overview}
    \end{figure*}

    Our previous work \citep{ashish2020} proposed an  approach for interpretable classification of Military Press exercises using videos as time series. We showed that a body pose estimation method, OpenPose \citep{openpose2019}, combined with multivariate time series classifiers (MTSC) can be used to accurately classify and interpret correct and incorrect executions. We henceforth name this approach BodyMTS (for Body tracking Multivariate Time Series). Figure \ref{fig:overview} shows the overall flow of BodyMTS: (1) pose estimation identifies and tracks multiple body parts over the video frames, (2) the $(X,Y)$ location coordinates of body parts for each frame are extracted resulting in multivariate time series, (3) a multivariate time series classifier is trained to classify the execution of the exercise into pre-defined classes. 
    %Our approach is a first step towards building a fully automated pipeline to classify the execution of physical exercises using video capture and light computation models suitable for resource constrained devices such as mobile phones. 
    
    In this paper, we extend our prior work with an extensive analysis of the robustness of BodyMTS to different sources of data noise, as well as a side-by-side comparison with state-of-the-art deep learning methods for human activity classification directly from videos.  Our hypothesis is that body pose estimation provides a strong prior for the classifier which now focuses on the pose information important for the task, not on other details in the video, e.g., background. This is contrasted to direct end-to-end video classification with deep learning, where noisy data may affect the model robustness and accuracy, and generalisation beyond benchmarks is known to be a challenge\footnote{\url{https://www.nature.com/articles/d41586-019-03013-5}} \citep{Azulay2019WhyDD}. In our experiments, we show that deep learning models require pre-training on large amounts of data, with a gap of 60\% in accuracy between training from scratch and pre-trained models.
    %The goal is to develop a lightweight, robust and faster application which can be easily executed on the resource constrained devices such as mobile phones. 
    Although BodyMTS in its current form is a proof-of-concept, we demonstrate its applicability and feasibility by considering the key factors that may influence performance, such as the impact of realistic noise types on the classifier accuracy and  running time, as well as the computational resources and storage space used by the data and models. We focus our attention on noise coming from changes in video quality, pose estimation quality, or time series data pre-processing. 
    %However, due to the unbounded sources of noises, we limit our focus on the noises coming from OpenPose, data pre-processing and recording conditions. 
    
    While research on assessing the performance of athletes using sensors has been successfully deployed\footnote{Two of the co-authors of this paper have successfully launched Output Sports, a start-up built on commercialising research results on single sensor systems: \url{https://www.outputsports.com}.}, there are currently not many approaches to classify the execution of strength and conditioning exercises using videos. 
    In our search, we have identified software such as Kinovea \citep{adnan2018biomechanical, puig2019validity} and DartFish \citep{fathallah2016evaluation, faro2016use}, which seem to work through manual analysis at a very low frame rate. Despite providing a vast number of features, these systems are not equipped with automatic classification of physical exercises \citep{adnan2018biomechanical, puig2019validity}.
    %Our approach, BodyMTS, aims at a fully automated pipeline to classify the execution of physical exercises using video data, with high accuracy, robustness to realistic sources of noise and satisfying the constraints set by the application, e.g., the machine learning models should not require extensive computational resources. 
    %Furthermore, recent works \citep{nakano2020evaluation, slembrouck2020multiview} observed that it is possible to obtain similar performance with marker less systems (video-camera + pose estimation) when compared with marker-based systems (expensive optical motion capture systems). 
    
    Existing research on human activity recognition from videos is based on applying complex deep learning architectures \citep{Shuiwang2010, Zisserman2014, Tran2015, Feichtenhofer2019}. Despite the competitive performance on benchmarks, this is achieved at the cost of heavy computation resources such as several hours of training and testing on high-end GPU hardware. Besides the need for high-end hardware, this also has a negative environmental effect. Furthermore, these models are trained and tested on datasets such as UCF-101 \citep{Soomro2012} and Kinetics-400 \citep{Kay2017}, which contain long duration videos and a wide range of activities. For instance, in Kinetics-400 the average duration of a clip is 10 seconds and the number of samples is around 300k. In our setting, a single clip is of 3 seconds duration on average and the differences between the classes are subtle, making the classification task more challenging, e.g., cycling vs walking in contrast to executing the MP with/without an arch in the back. Our dataset is also small (a few thousand samples for training and validation) when compared to these large benchmarks. We have found no prior work that uses videos for strength and conditioning exercise classification and works with this type of smaller data scale and fine-grained classification.
    
    Our main contributions in this paper can be summarized as follows:
    \begin{itemize}
    %\item %We present the updated results using the latest version of OpenPose (v1.7). OpenPose v1.7 is way faster and efficient than the OpenPose v1.4 and tracks 25 body parts as compared 18 body parts in OpenPose v1.4. Additionally, we measure the classifier accuracy for different subsets of dimensions of the data. We further include another recent state-of-the-art multivariate time series classifier MINIROCKET for comparison. 
    \item We present and extensively evaluate BodyMTS, an end-to-end video-as-timeseries human exercise performance classification method. We study the impact of improvements in body pose estimation methods (e.g., OpenPose, \citep{openpose2019}) and recent multivariate time series classifiers (e.g., ROCKET \cite{Dempster2020} and MiniROCKET \citep{Dempster2021MINIROCKETAV}) on the overall classification accuracy. We show improvements in accuracy with an average of 87\% classification accuracy for the Military Press exercise.
    
    \item We analyze the robustness of BodyMTS against different types of realistic noise and measure the impact on the classifier performance. We consider three common sources of noise in our application setting: video capture quality, pose estimation quality and time series pre-processing steps.
    %We quantify the amount of noise present using standard scores such as VMAF.

    \item  We conduct an extensive empirical study comparing BodyMTS to state-of-the-art deep learning approaches for human activity recognition directly from videos. We compare all methods in terms of accuracy, training/testing time and computation resources. 
    %We further investigate the impact of noise on accuracy. 
    We show that BodyMTS is robust to lower quality data captured at prediction time and has fast training and prediction.
%    requires much less computation resources to achieve the same accuracy.
    %with much less computational resources and running time.
    
    \item To support our paper, all of our code, data and detailed results are available at:  \url{https://github.com/mlgig/BodyMTS_2021.git}.
    
    \end{itemize}
    
    The paper is organized as follows.  In Section \ref{sec:app_require} we discuss the application and technical requirements of BodyMTS. In Section \ref{sec:relwork}, we give an overview of the related literature on human activity recognition, human pose estimation, strength and conditioning exercises and multivariate time series classification. Section \ref{sec:data} describes the data collection process and the Military Press  dataset.  
    Section \ref{sec:methods}, presents our methodology for classifying MP exercises from videos and Section \ref{sec:dmc} describes the main data mining challenges. In Section \ref{sec:exp}, we analyze the robustness of BodyMTS against different sources of noise and compare its performance with state-of-the-art deep learning methods. In Section \ref{sec:discussion}, we describe the lessons learned from this study, as well as limitations and future work. In Section \ref{sec:recomm_prac} we summarise our recommendations for practitioners working on similar tasks and 
    we conclude in Section \ref{sec:conclusion}.

\section{Application Requirements}
\label{sec:app_require}

In this section, we discuss the required BodyMTS features and the corresponding application and technical requirements. 
%We also discuss the working of each component involved in BodyMTS. 
We note that BodyMTS is currently a proof-of-concept and the actual deployment scenario and requirements may change depending upon the business case and the end-user requirements.

The aim of BodyMTS is to provide a scalable system that can accurately measure and evaluate end-user performance of strength and conditioning (S\&C) exercises, with a view to provide feedback in near real-time. This, in turn can guide physiotherapists, trainers, and elite and recreational athletes to perform exercises correctly and therefore minimise injury risk and enhance performance. We devised the following list of application requirements based on previous research in which we consulted with end users, clinicians, and strength and conditioning experts on the design, implementation and evaluation of interactive feedback systems for exercise \citep{Brennan2020,Argent2019, Argent2018,Giggins2015,Martin2017}:

\begin{itemize}
    \item Be able to accurately monitor the body parts movement, accounting for the critical body segments involved in the exercise in question.
    
    %This is determined by calculating the minimally detectable change (MDC). However, this level of MDC will be different for different exercises so an overall benchmark cannot be determined. Hence, in this case we chose the best pose estimation method which can accurately track the body pose segments by evaluating based on their performance on benchmark datasets. 
    
    \item Detect when deviations from normal movement profile have occurred, and which kind of deviation has occurred in each case.
    %Previous research \cite{whelan2019determining} has demonstrated very poor intra and inter-rater reliability for domain expert identification of movement deviations in commonly performed multi-segment screening exercises. The minimum requirement in this case would be to outperform this state of the art, but this would not be very challenging. Kappa scores in the study by  \citep{whelan2019determining} at an inter-rater level ranged from 0.18 to 0.53, and intra-rater agreement ranged from 0.38 to 0.62. Therefore, establishing a threshold of 80\% classification accuracy would be a reasonable benchmark for success in this system. This would mean that 8 out of every 10 executions are correctly classified in terms of identifying whether it has been performed to an acceptable level, and which specific deviation has occurred in the case of aberrant performance.
    
    \item Provide clear and simple feedback to the end user, in near real-time.
    %The end user must receive feedback on the performance of a set of exercise repetitions immediately after the set has been completed. This will require near to real time capture, processing and analysis of the data. The feedback should be meaningful, in the sense that it should give an explanation as to which part of the movement, and which body locations, are responsible for the observed deviation. Doing so will enable the end-user to address this deviation in the subsequent performance of the exercise
    
    \item Simple data capture based on ubiquitous sensor technology (e.g, single phone). 
    %Ideally, the system should be based on a sensor technology that is very inexpensive, or at the least is likely to be already owned by the end user. The system should be lightweight, easily deployable, and not require extensive setup and calibration prior to each use.
    
    \item Coverage of wide range of S\&C or rehabilitation exercises. 
    %The system should be scalable and transferable, in the sense that coverage for additional exercises does not involve an exhaustive data capture and model training process.
\end{itemize}
Table \ref{table:body_requirements} summarizes the application features and the corresponding application and technical requirements for such a system.

\begin{table}[hbt!]
  \small
\centering
\begin{tabular}{ p{3cm} | p{4cm} | p{3.5cm}}
    \textbf{Application Features} & \textbf{Application Requirements} & \textbf{Associated Technical Requirements} \\
    \hline
    % Pose estimation
    Accurate real-time measurement of body part movements involved in the exercise.
    &
    % Level of accuracy and reliability for tracking anatomical landmarks or estimation of joint angles that will facilitate clinically relevant levels of minimally detectable change (MDC) for specific exercises. This level of MDC will be different for different exercises so an overall benchmark cannot be determined.
    
    Level of accuracy and reliability for tracking anatomical landmarks or estimation of joint angles that will facilitate detection of clinically relevant changes in performance of specific exercises.
    %known as Minimal Important Difference. 
    This will differ across exercises and contexts (e.g. greater level of sensitivity required in elite athletes compared to recreational athletes) so an overall benchmark cannot be specified.  
    
    &
    %\begin{itemize}
    Accurate real-time human pose estimation for the body parts relevant to the specific exercise:
    OpenPose \citep{openpose2019} can detect 25 body parts in an input image in under 1 second and has accuracy above 75\% on established benchmarks.
    %\item Performance on benchmark datasets.
    %\end{itemize} 
    \\
    % Detect deviations
    System can detect when deviations from perfect movement profile have occurred, and which kind of deviation has occurred in each case. 
    & 
    Previous research \cite{whelan2019determining} has demonstrated poor intra and inter-rater reliability for domain expert identification of movement deviations in commonly performed screening exercises. %The minimum requirement in this case would be to outperform this state of the art. 
    Kappa scores at an inter-rater level ranged in 0.18 - 0.53, and intra-rater agreement ranged in 0.38 - 0.62. Therefore, establishing a threshold of 0.8 classification accuracy would be a reasonable benchmark for success in this system (i.e., 8 out of every 10 executions are correctly classified). 
    %in terms of identifying whether it has been performed to an acceptable level, and which specific deviation has occurred in the case of aberrant performance.
  
    &
    Accurate and efficient classification:
    ROCKET \citep{rocket,dhariyal20mtsc, ruiz20mtsc} has state-of-the-art accuracy on established MTSC benchmarks  and is a very efficient classifier. In our experiments it achieves a classification accuracy higher than 80\%.
    %\item Comparison with other existing methods.
    \\
    % Feedback
    Clear and understandable feedback to end user at an appropriate time (near real-time).
    
    & 
    The end-user must receive feedback on the performance of a set of exercise repetitions immediately after the set has been completed. This will require near real-time capture, processing and analysis of the data.
    The feedback should be meaningful: it should give an explanation as to which part of the movement, and which body locations, are responsible for the observed deviation; should enable the end-user to address this deviation in the subsequent performance of the exercise.
    
    &
    Class label feedback: Near real-time class prediction on single video clip. 
    BodyMTS completes data processing and prediction in 25 seconds for a clip of 10 exercise repetitions.
    Simple and accurate explanation: 
    %Explanation can be provided through saliency maps on MTSC data, eg with methods such as ROCKET-LIME \citep{surabhi:aaltd21}. 
    Explanation is beyond the scope of this paper.
    \\
    % Sensor and video
    Simple data capture, with ubiquitous sensor technology.
    
    & 
    Ideally, the system should be based on  sensor technology that is very inexpensive, or at least is likely to be already owned by the end user.
    The system should be easily deployable, and not require extensive setup and calibration prior to each use.
    
    &
    Video data capture through mobile phone.
    Data reduction to enable fast processing, while maintaining accuracy: BodyMTS can reduce the data size by 70\% and still preserve classification accuracy higher than 80\%.
     \\
    % Wide range of activities
    Coverage for wide range of S\&C or rehabilitation exercises.
    
    & 
    The system should be scalable and transferable, in the sense that coverage for additional exercises does not involve an exhaustive data/feature engineering and model training process.
    & 
    Fully automated pipeline: BodyMTS is fully automated.
    Inclusion of other exercises: This is beyond the scope of this paper.
    \\
    
    % Recording Conditions
    % Recording Conditions.
    % & 
    % \begin{itemize}
    % \item Centralize the participant.
    % \item Stable camera places on a horizontal surface.
    % \item Viewing angle for each exercise.
    % \end{itemize}
    % & 
    % \begin{itemize}
    % \item Mobile application that take cares of the bounding box.
    % \end{itemize} \\
    
    \bottomrule
\end{tabular}
\caption{Application features, application requirements and  associated technical requirements.}
    \label{table:body_requirements}
\end{table}
    
%In the next section, we discuss the components and workflow of BodyMTS. 
There are two main components of BodyMTS:
\begin{itemize}
    \item \textbf{Client side mobile application} which the end-user uses to record their execution. The recorded videos are then pre-processed before sending them to the server side running on the cloud.
    \item \textbf{Server side application} which stores the pre-trained model. Each repetition of the clip is classified separately using the stored model. The final results are then returned back to the client side mobile application. 
\end{itemize}

Figure \ref{fig:overview} shows the overall workflow of BodyMTS. Before the user starts using the client side mobile component of BodyMTS, we expect the following requirements to be fulfilled:

\begin{itemize}
    \item The mobile camera used for recording the execution should be placed on a static surface before the start of the execution. 
    
    \item The view of the camera will vary depending upon the type of exercise (e.g., front view for Military Press). 
    %As we show in the Appendix, the front view works best for Military Press. 
    This is static information that will be already stored within the application for each type of exercise.
    %the user will place camera as mentioned for each type.
    
    \item The mobile application will use a bounding box to centralize the user with respect to each frame. 
\end{itemize}
These are quality control requirements that are evaluated before and during the execution. The video and pose estimation have to be of sufficient quality, otherwise the data will be rejected by the application.
After the above conditions are fulfilled, the user will activate the application and start recording the video using the mobile application. At the end of the workout, the client side application will pre-process the data. The recorded video will be pre-processed to centralize the participant and to remove the audio. Further, it will be compressed to reduce the total size and will be sent to the server side application where further processing takes place. 
At the server side, the compressed videos undergo pose estimation followed by segmentation and classification steps. Finally, the classification results are returned back to the client side mobile application. 
\clearpage
%The frequency of re-training the stored pre-trained model will be determined based on factors such as personalized training program and computing resources.     
\section{Related Work}
\label{sec:relwork}

In this section we present an overview of existing approaches for strength and conditioning exercise classification,  human action recognition from videos, human pose estimation and multivariate time series classification. %Additionally, we provide a background towards the existing methods to handle noise in the data. 
% We present an overview of existing approaches for human activity recognition from video, and discuss the latest advances in pose estimation and multivariate time series classification methods and their interpretation.

\subsection{Strength and Conditioning Exercise Classification}

The purpose of S\&C exercises is to improve the performance of athletes in terms of strength, speed, flexibility, agility \citep{kinectyoga, Martin2018, chu2019artificial, Martin2015}. S\&C exercises span multiple types of exercises or movement sequences that target different parts of the body and different functional goals. In some cases, the person interacts with a weight or mechanical apparatus, whereas in others the person performs a free body movement without any interaction with an external system or force (e.g., jump). Recent advances in technology have spurred the usage of high tech solutions to maximize the performance of athletes. 
These can be divided into three broad categories: optical motion capture, wearable inertial sensors and video \citep{ashish2020, puig2019validity, faro2016use, fathallah2016evaluation}. 

The most popular optical motion capture system is Microsoft Kinect. The work of \citep{kinectyoga, zerpa2015use, ressman2020reliability, decroos0BVD18, DAJIME2020104021} has investigated the use of Microsoft Kinect for rehabilitation exercises, movement quality assessment and gait analysis. However, despite their high performance, these systems are expensive, need high maintenance, require significant time to set up and are mostly limited to controlled clinical trials. 
% Moreover, the data extracted from these devices has to undergo manual feature engineering which in turn requires domain expertise \citep{decroos0BVD18, DAJIME2020104021}. 
% Additionally, these systems lack evidence of reliability and validity for human kinematics analysis \citep{zerpa2015use, DAJIME2020104021}. is not well suited for lower limb exercises \citep{decroos0BVD18}

Wearable inertial sensors-based approaches consist of fitting Inertial Measurement Units (IMU) \citep{Martin2018, chu2019artificial, espinosa2015inertial} on different parts of the body. The sensor data is analyzed to evaluate performance using supervised machine learning methods, visualization or manual techniques. The number of inertial sensors required and their positions vary from exercise to exercise \citep{espinosa2015inertial, Darragh2016, Martin2018, Martin2017}. Research methods and also commercial systems have been deployed using such inertial sensors.
Still, sensors can be expensive, they may hinder the ease of movement particularly when applied over many body parts and over longer periods of time, and the annotation process can be time-consuming \citep{Darragh2016, Kwon2020, DAJIME2020104021}. 

The third category uses video-based devices such as dedicated cameras (DSLR) or smartphone cameras to capture data. Proprietary software such as Dartfish \citep{fathallah2016evaluation, faro2016use} and open-source software such as Kinovea \citep{adnan2018biomechanical, puig2019validity, moral2015agreement} are used to analyze performance by providing the option of slow-motion replay at a very low frame rate. However, these systems are less accurate and require fitting body markers of different color to the background. Recent work \citep{Slembrouck2020, Nobuyasu2020, Stamm2020} utilizing pose estimation for motion tracking has paved the way for alternative approaches to IMUs and optical motion capture systems. We found no prior work that utilizes video to classify S\&C exercises. 
%and avoid manual feature engineering.
%We did not find 
% Our work differ from other approaches in two ways: we directly utilize the raw data from videos without manual feature engineering using multivariate time series classification
% Most research methods focus on deep learning methods \citep{Feichtenhofer2019, Carreira2017, Tran2015} for human activity recognition directly from videos. 

\subsection{Human Activity Recognition}

Video-based Human Activity Recognition (HAR) is a core area of computer vision. 
%facing challenges such as high data dimensionality, viewpoint variations, background cluttering and complex motion dynamics.
HAR methods can be broadly classified into two categories. First are methods based on handcrafted features such as bags of visual words \citep{Heng2013, Dalal2006, Peng2014, Sanchez2013}. These include finding local spatio-temporal features such as motion boundary histograms \citep{Dalal2006} and trajectories \citep{Heng2013}, followed by feeding them to a classifier. These methods have been shown to provide competitive performance on benchmark datasets \citep{Carreira2018, Sigurdsson2016, Soomro2012} before the emergence of deep learning methods. 
The second category includes deep learning methods, in particular convolutional neural networks. The recent success of 2D-CNN \citep{Krizhevsky2012} in image classification has motivated researchers to employ these models for action recognition in video. Several models, e.g., 3D-CNN \citep{Shuiwang2010}, two stream convolutional networks \citep{Zisserman2014}, I3D \citep{Carreira2017} and Slowfast \citep{Feichtenhofer2019} have achieved state-of-the-art performance on benchmark datasets. These models are  computationally expensive, we found no studies evaluating these methods for strength and conditioning exercise classification, under specific application constraints. It is also not clear how well these methods work on real use cases.
%Approaches based on representing videos as space-time regions \citep{Wang2018} and using temporal layers \citep{Hussein2019} are able to overcome these limitations. Other architectures employ attention-based mechanisms \citep{Gir2019, Zhenyang2018, Shar2015} to focus on important regions of the video. 

\subsection{Human Pose Estimation}
Pose estimation refers to recognizing the postures of humans by detecting the body parts from images. It is considered one of the hardest problems in computer vision due to challenges such as occlusion, complex motion dynamics, interactions and background \citep{openpose2019,Papandreou2017,Huang2017}. 
%The task becomes even more difficult when dealing with multiple persons in the same image or video frame. 
%Pose estimation has many applications, such as video surveillance and action recognition. 
Traditional approaches \citep{Andriluka2009, Gkioxari2013, DSapp2013, Dantone2013} were based on extracting handcrafted features. Current methods based on deep learning architectures \citep{openpose2019, Papandreou2017, He2017, Newell2017} have achieved remarkable results on this task. 
%These architectures can be classified into two categories: single-stage and multi-stage. Single stage networks employ pre-trained image classification models based on VGG16 \citep{VGG16} or ResNet \citep{Resnet}.
%https://arxiv.org/pdf/1711.07319.pdf
%ingle stage networks consists of architectures such as Mask R-CNN \cite{He2017} which uses Region Proposal Network (RPN) to generate the region proposals, and Cascade Pyramid Network (CPN) \cite{Yu2018} which uses global pyramid network (GlobalNet) and pyramid refined network (RefineNet). Multi-stage architectures can be further divided into sub-categories: top-down and bottom-up. Top-down \cite{Papandreou2017, Huang2017, He2017} approaches use off-the-shelf person detector over the whole image, which is followed by detecting the joints individually, whereas more recent bottom-up approaches first focuses on detecting the joints which is followed by assigning them to the corresponding person. 
%\textbf{Top-down} approaches include methods such as Mask R-CNN \citep{He2017} which employs Region Proposal Network (RPN), to generate the region proposals or candidates which is RoI pooled according to the region, and goes through the remaining network. The work based on \cite{Papandreou2017} uses the Faster RCNN detector to predict the location and scale of boxes followed by estimating the keypoints of the person potentially contained in each proposed bounding box. 
Recent approaches include methods such as OpenPose \citep{Newell2017, Insafutdinov2016, openpose2019}, which work by first finding the body joints and associating them using affinity fields, and DeepCut \citep{Pishchulin2015}, which uses a partitioning and labeling formulation of a set of body-part hypotheses generated with CNN-based part detectors. 
%Other architectures based on deep learning \citep{Zhou2016, Chen2018} and hourglass architectures \citep{Newell2016} have been proposed and have been shown to give competitive performance on  benchmarks.
%Top-down approaches are slower than bottom-up approaches \cite{openpose2019} in case of multi-person pose estimation as top-down employs person detector for every single person. However, top-down approaches can outperform bottom-up approaches in terms of accuracy. 
OpenPose can detect and track multiple body points in real-time and with high accuracy. The most recent version of OpenPose \citep{openpose2019} can detect 25 body parts in an input image in under 1 second with average accuracy ranging from 75.6\% to 79\% on recent 2D pose estimation benchmarks. 
%While the output of pose estimation methods is still noisy, we model this output with the latest time series classification methods, which are designed to be more robust to noise.

\subsection{Multivariate Time Series Classification}

Time series classification is a form of supervised classification where the data is ordered. 
%Each sample has a single dimension and a class label in case of univariate, whereas each 
For multivariate time series classification (MTSC), each sample has multiple dimensions and a class label. 
%Until recently, the research in this area focused on classifying univariate data. However, it is more common to have multivariate data where multiple dimensions are being generated at the same time. 
%For instance, there are multiple sensors that track the body motion in human activity recognition, such as acceleration and gyroscope. 
%Most of the methods for MTSC are based on extending UTSC techniques. However, this is not effective and most of them suffer from long execution time and consuming too much memory. It is thus important to develop effective multivariate time series classification (MTSC) methods. 
We can group existing methods for MTSC into five broad categories \citep{DBLP:journals/datamine/RuizFLMB21, dhariyal20mtsc}: distance-based, feature-based, ensemble based, linear models and deep learning. These methods have mostly been evaluated on the UEA MTSC \citep{timeseriesclassification} archive, which contains 30 multivariate datasets. Among the methods evaluated, linear classifiers and deep learning methods have achieved high accuracy, with low running time and excellent scalability, hence we focus on this subset here.

\noindent\textbf{Linear Classifiers}. ROCKET \citep{dempster2019rocket} (RandOm Convolutional KErnel Transform) is the current state-of-the-art for both univariate and multivariate TSC in terms of accuracy and scalability. It uses a large number of random convolutional kernels in conjunction with a linear classifier. MINIROCKET \citep{Dempster2021MINIROCKETAV} is a recent extension of ROCKET. It is deterministic, faster and more efficient than ROCKET. %It utilizes a smaller set of kernels along with exploiting the various properties of kernels as compared to ROCKET. 
Unlike ROCKET, MINIROCKET implicitly normalizes the time series, thus it is scale invariant, which for our application proves to be a weakness; for MP the magnitude of the signal plays an important role in the classification task and thus the signal should not be normalized.

\noindent\textbf{Deep Learning Classifiers.} Recent success on image classification \citep{VGG16, krizhevsky2012imagenet} has motivated researchers to use deep learning methods to classify time series data. %However, the application of deep learning based methods for classifying the multivariate data is still at a very early stage. 
\cite{Fawaz2019} presented a comprehensive study of 9 deep learning models to classify univariate and multivariate time series. Fully Convolutional Networks (FCN) and ResNet have shown state-of-the-art performance  without suffering from high time and memory complexity. 

\section{Data Collection}
\label{sec:data}

\begin{figure}
    \centering
    \includegraphics[scale=0.8]{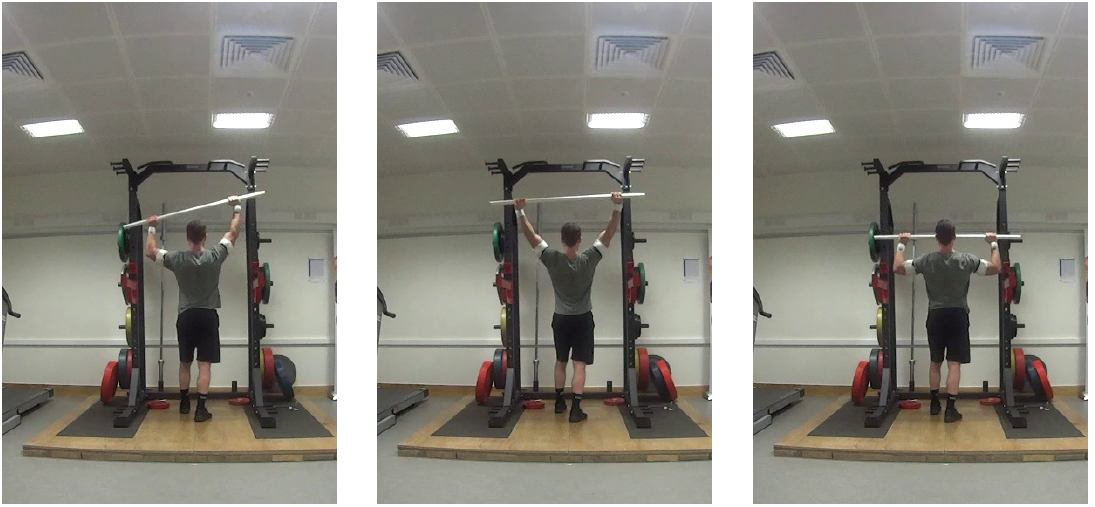}
    \caption{Single frames depicting the induced MP deviations for class A, Arch and R (left to right).}
    \label{fig:deviations}
    \end{figure}

    \textbf{Crossfit Workout Dataset.}
    %We have considered the execution of the Military Press exercise for the current work. 
    The data used for evaluating our approach consists of video recordings of the execution of the Military Press exercise (MP). During this exercise the barbell is lifted to shoulder height and then smoothly lifted overhead by extending the elbows. The amount of weights lifted and time taken for each repetition may vary from participant to participant. MP is an important exercise in strength and conditioning, injury risk screening, and rehabilitation \citep{Darragh2016}. 
    %Incorrect execution of MP may lead to muscoskeletal injuries and impede performance \citep{Baechle2008}.
    Participants were asked to complete fixed repetitions of normal and aberrant forms for this exercise.
    %The induced forms refer to predefined deviations as defined by the National Strength and Conditioning Association (NSCA) \cite{Baechle2008}. 
    Figure \ref{fig:deviations} shows some examples for the execution of the MP exercise.
        
    \textbf{Participants.}  
    53 healthy volunteers (31 males and 22 females, age: 26 $\pm$ 5 years, height: 1.73 $\pm$ 0.09 m, body mass: 72 $\pm$ 15 kg) were recruited for the study. 
    Participants did not have a current or recent musculoskeletal injury that would impair performance of multi-joint upper limb exercises. The Human Research Ethics Committee at University College Dublin approved the study protocol and written informed consent was obtained from all participants before the study start.
	
	\textbf{Experiment Protocol.}
	The testing protocol was explained to participants upon their arrival at the laboratory. Participants completed 10 repetitions of the normal form and 10 repetitions with induced deviations. In order to ensure standardization, the technique was considered acceptable if it was completed as defined by the National Strength and Conditioning Association (NSCA) guidelines. The induced forms were chosen based on common deviations listed in the NSCA guidelines \citep{Baechle2008} and through discussion with sports physiotherapists and strength and conditioning coaches. 
	Participants were allowed to familiarize themselves by completing practice repetitions. 
	%No real weights were used during the execution and all the participants were very fit and healthy. 
	All the performances were observed and labelled by an expert. If the performance had degraded due to fatigue then this would have resulted in the data being excluded at source. Each repetition was observed and if any repetition was not consistent with the label (based on domain expert observation) it would have been excluded from the data.
	
	Two cameras (Sony Action Camera, Sony, Tokyo, Japan) were set up in front and to the side of the participants to allow for recording in the frontal and lateral planes simultaneously. The data is recorded at a rate of 30 frames per second with 720p resolution. Each of these individual video clips were then labeled according to participant number, exercise completed and if they were completed in an acceptable or aberrant manner. Each participant completed the set at their desired tempo. 
	
	% Exercise technique and deviations
    \textbf{Exercise Technique and Deviations.}
    The induced forms were further sub-categorized depending upon the exercise. The deviated execution forms undermine the performance of the participants leading to a higher chance of injury. Completing the exercises with this aberrant technique means strength gains are not made as efficiently and can increase the likelihood of injury. Below we describe the four classes of normal and deviated execution forms for the MP exercise.
    
    \textbf{Exercise Classes. Normal (N):} This class refers to the correct execution of the exercise. The participant starts by lifting the bar from near shoulder to all the way above the head until the arms are fully stretched and then bringing it back to shoulder level with no arch in the back. The bar must be stable and parallel to the ground and the back should be straight.
    \textbf{Asymmetrical (A):} This form refers to the execution when the bar is lopsided and asymmetrical.
    \textbf{Reduced Range (R):} This form refers to the execution when the bar is not brought down completely to the shoulder level.
    \textbf{Arch (Arch):} This type of execution indicates that the participant arches their back. 
    
\section{Methods}
\label{sec:methods}

    In this section we present the BodyMTS pipeline and provide details about individual components and data pre-processing steps. We give a description of OpenPose and why it is considered above other pose estimation libraries. We also briefly explain the process of obtaining multivariate time series data from videos using OpenPose, segmenting the long time series to obtain individual time series for each repetition and the methods chosen for the classification.

    \subsection{Methodology}
    \label{sec:proposed_approach}

    % \noindent\textbf{Proposed Approach.}
    BodyMTS is a novel end-to-end approach to classify video-based S\&C exercises. It consists of two main steps. The first step applies  human pose estimation to extract multivariate time series data from video. The time series data is obtained by applying pose estimation, which tracks the location coordinates of multiple body parts over the video frames. The second step applies multivariate time series classification methods. 
    %We discuss next the details of body pose estimation, time series classification as well as data pre-processing steps that are important for this methodology.
    
    %%%%%%%%%%%%%%%%%%%%%%%%%%Start Pose Estimation Explanation%%%%%%%%%%%%%%%

    \noindent\textbf{Body Pose Estimation.} We select 
    OpenPose \citep{openpose2019} over other frameworks such as R-CNN \citep{RCNN} or Alpha-Pose \citep{alphapose}, for the following reasons: (1) it is robust against possible occlusions including during human-object interaction; (2) it is a full-fledged system and does not require manual steps such as generating all frames for videos, setting up a display to visualize the results and saving the results in a desired format; (3) it can run on different platforms, including Ubuntu, Windows, Mac OSX, and embedded systems; (4) the inference time of OpenPose outperforms all state-of-the-art methods, while preserving high-quality results. OpenPose also provides a confidence score for each body part and each frame. The confidence score ranges from 0.0 to 1.0, where a higher confidence value indicates a higher probability of detecting a particular body part in that location. 
    This score can be utilized as a proxy to assess the accuracy of OpenPose. 
    %The higher the confidence score, the better is the accuracy of OpenPose. This score is averaged for every body part over a single clip to calculate the OpenPose confidence of a single clip. This is used to compare the accuracy of OpenPose under different sources of noise.

    \iffalse
	\begin{figure}[h!]
    \centering
    \includegraphics[width=4cm,height=7cm]{figs/keypoints_pose_25.png}
    \caption{OpenPose tracks 25 points on the human body. Figure taken from
    OpenPose website.}
    \label{fig:keypoints_human_body}
    \end{figure}
    \fi 
    
    \begin{table}[h!]
    \centering
    \resizebox{0.7\columnwidth}{!}{%
    \begin{tabular}{lll}
    \toprule
        0 Nose              &  8 Mid Hip                 &  16 Left Eye \\
        1 Neck              &  9 Right Hip               & 17 Right Ear \\
        2 Right Shoulder    &  10 Right Knee             & 18 Left Ear   \\
        3 Right Elbow       &  11 Right Ankle            & 19 Left Big Toe \\
        4 Right Wrist       &  12 Left Hip               & 20 Left Small Toe \\
        5 Left Shoulder     &  13 Left Knee              & 21 Left Heel  \\ 
        6 Left Elbow        &  14 Left Ankle             & 22 Right Big Toe \\
        7 Left Wrist        &  15 Right Eye              & 23 Right Small Toe \\   
        & & 24 Right Heel  \\   
        \bottomrule
    \end{tabular}
    }
    \caption{The 25 body parts tracked by OpenPose (v1.7) in a video frame \citep{openpose2019}.}
    \label{table:all_body_parts}
    \end{table}
    
    Table \ref{table:all_body_parts} shows the list of body parts tracked by OpenPose. It outputs the $X$, $Y$ coordinates and detection confidence, for the detected body parts in a given input frame. 
    % Frames are first cropped to remove unnecessary background and center the participant in the frame. Cropping involves removing the extreme left and right portion of the image. This step does not risk cropping the participant as the camera settings remain unchanged for all participants. 
    The coordinates are obtained with origin as a reference at the top left side of the image. The video is fed to OpenPose to obtain a sequence of $X$ and $Y$ location coordinates for each body part and each video frame. 
     
\begin{figure}
    \centering
    \includegraphics[width=0.95\columnwidth]{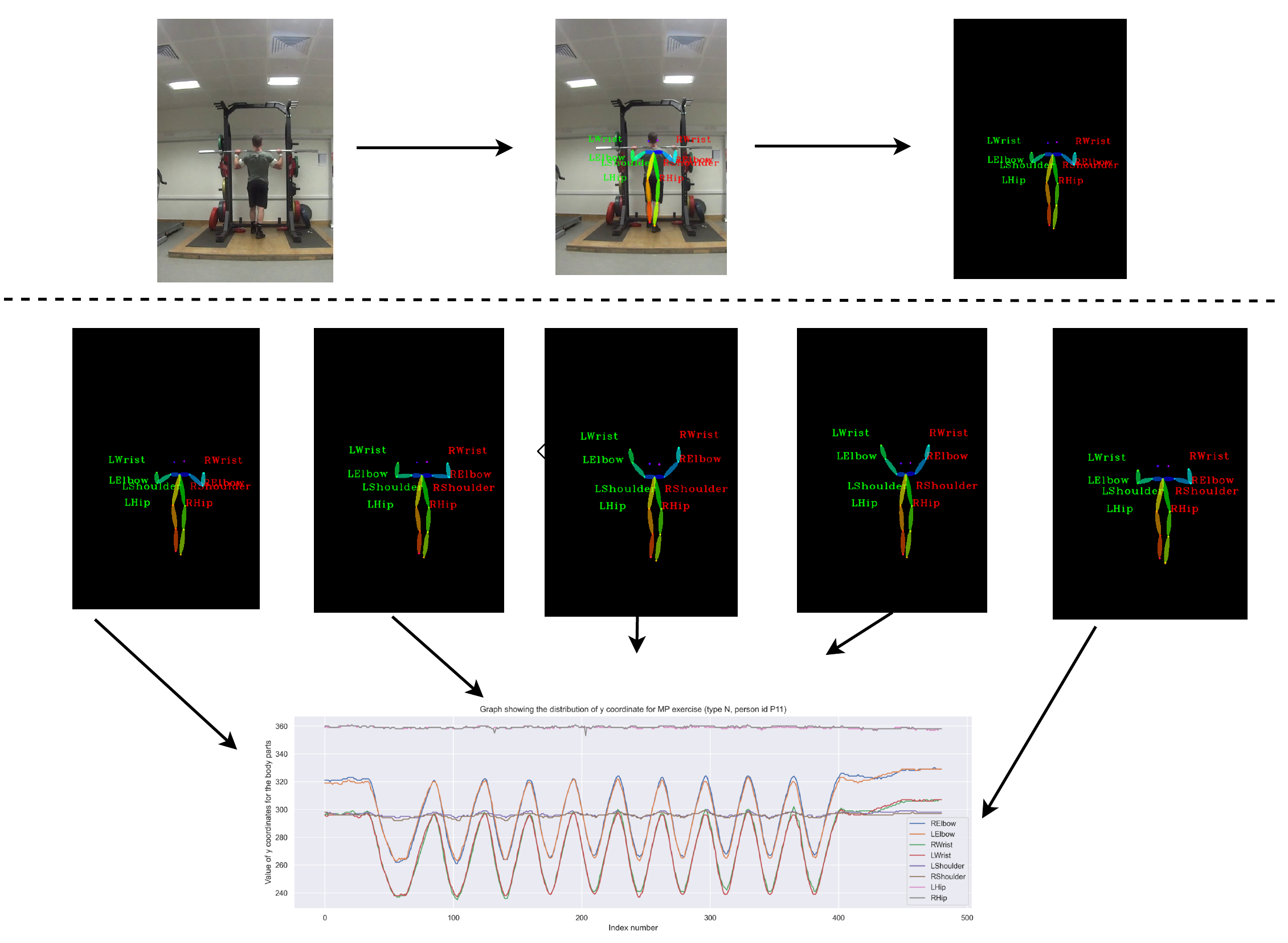}
    \caption{Extraction of time series data from video using OpenPose. Each frame in the video is considered as a single time point in the resulting time series. Each tracked body part results in a single time series that captures the movement of that body part. The whole motion is captured as a multivariate time series with 50 channels, two (X,Y) channels for each body part tracked (only 8 body parts with Y coordinate shown above). A  class label is associated with each such multivariate time series.}
    \label{fig:videotots}
    \end{figure}

    Each frame is considered a single time point in the output time series data. The original videos require 813MB of storage and after cropping and removing the audio this reduces to 213MB. After applying OpenPose and extracting the body parts time series, the data size reduces to 30 MB, roughly a reduction of 7 times. Figure \ref{fig:videotots} shows the use of pose estimation to track the coordinates of body parts over all the video frames. The plot at the end shows the raw $Y$-coordinates for 8 upper body parts for a single video from class $N$. The time series obtained for lower body parts such as ankles, hips, etc., do not show much variability throughout the whole clip as these body parts mostly remain static throughout the execution of the Military Press.
    
    % https://learnopencv.com/multi-person-pose-estimation-in-opencv-using-openpose/
    
    %  # cropped 24.88; original 64.78 seconds
    %  # 720 * 1280 --> 576 * 384
    
    % GI : make the figure bigger

%%%%%%%%%%%%%%%%%%%%%End Pose Estimation Explanation%%%%%%%%%%%%%%%%%%%

%%%%%%%%%%%%%%%%%%%%%%%%%%%Start Pre-processing%%%%%%%%%%%%%%%%%%%%%%%%%%%%%%
%%%%%%%%%%%%%%%%%%%%%%%%%%%%%%%%%%%%%%%%%%%%%%%%%%%%%%%%%%%%%%%%%%%%%%%%
    \noindent\textbf{Multivariate Time Series Data.}
    Each video records 10 repetitions for each exercise execution, resulting in a time series capturing the body points movement for 10 repetitions. Since each repetition is the record for a single exercise execution, segmentation of the long time series is required to obtain the sequence for a single repetition. Each repetition forms a single time series sample for training and evaluating a classifier. We use peak detection methods to segment the pose estimation time series data. We find the location of local maxima in the signal using the scipy package\footnote{\url{https://github.com/scipy/scipy}}. The body parts that are considered for finding the peaks are elbows or wrists, as these are the only body parts showing regularity in the patterns as shown in Figure~\ref{fig:videotots}.  We only keep the upper body parts time series as suggested by the domain experts who carried out the data collection. 
    %The next step consists of dropping the body parts time series with low variability such as ankles, knees, toes etc.
    % Additionally, the sequences of $X$-coordinates are ignored due to low variability. 
    The data for some body parts (nose, eyes) are ignored as OpenPose fails to track these  since the participant is not facing the camera. We also present results with different subsets of dimensions to understand the impact of different body parts on accuracy. We investigate using all the channels or using automated channel selection methods (Section \ref{sec:exp}).

    The time series obtained after this step have variable length since the time taken to complete each repetition differs from participant to participant. Since the current implementation of ROCKET and deep learning  methods cannot handle variable-length time series, all time series have been re-sampled to the length of the longest time series (with a length of 161). We use a 1D interpolation function with a cubic spline fit to interpolate each time series to same length \citep{2020SciPy-NMeth}.
    
    The final data contains time series corresponding to 8 body parts (elbows, shoulders, wrists and hips) with 16 channels (X and Y coordinates). Lastly, as observed in our prior work \citep{ashish2020} the time series data is not normalized as normalization leads to a substantial drop in accuracy. Through this application we learned that most state-of-the-art time series classifiers only work with fixed-length time series and also have an implicit step of normalizing the time series. While these algorithmic constraints seem harmless for clean TSC benchmarks, they prove problematic for real use cases. Re-sampling the length changes the meaning of the time series, and similarly, default or implicit normalization done within the algorithm, changes the meaning of the data and affects the accuracy of the classifier. 
    %In this regard, we invite the TSC community to carefully consider these constraints in the classifier design, try to design more flexible algorithms and expose some of these choices to the user. 
    Among the classifiers evaluated, only ROCKET exposes the option of data  normalization to the user, and this makes a 10 percentage points difference in accuracy for our application.

%%%%%%%%%%%%%%%%%%%%%%Start Train/Test Split%%%%%%%%%%%%%%%%%%%%%%%%%%%%%%%
%%%%%%%%%%%%%%%%%%%%%%%%%%%%%%%%%%%%%%%%%%%%%%%%%%%%%%%%%%%%%%%%%%%%%%%% 

    \noindent\textbf{Train/Test Split.}
    We perform repeated 70:30 splits on the full data set to obtain training and test data. Each split is done based on the unique participant IDs to avoid leaking information into the test data. By splitting on the ID level we make sure that all the samples from a particular participant go into either the training or test data. The data is overall balanced across all the classes. Table \ref{table:train_test_split_stats} shows the number of samples across the four classes, for a single train/test split. There are roughly 1400 and 600 samples in training and test data respectively.
    
    \begin{table}[ht]
    \centering
    \resizebox{0.45\columnwidth}{!}{%
    \begin{tabular}{l|l|l|l}
        \textbf{Class} & \textbf{Training} & \textbf{Test} & \textbf{Total} \\
        \hline
        N & 370 & 150 & 520 \\
        A & 360 & 150 & 510 \\
        R & 361 & 151 & 512 \\
        Arch & 361 & 150 & 511 \\
        \textbf{Total} & \textbf{1452} & \textbf{601} & \textbf{2053} \\
    \end{tabular}
    }
    \caption{Total number of samples per class in the train and test datasets for one 70:30 split.}
    \label{table:train_test_split_stats}
    \end{table}
    
%%%%%%%%%%%%%%%%%%%%%%%End Train/Test Split%%%%%%%%%%%%%%%%%%%%%%%%%%% 

    % \textbf{Definition}: A multivariate time series with $M$ dimensions can be defined as
    % $X = [X^1, X^2,...X^M]$ 
    % where each $X^i$ is a univariate time series with $X^i$ $\epsilon$ $R^T$ where $T$ is the length of the time series. In our case, the value of $M$ is 8 (corresponding to 8 body parts) and the value of $T$ is 161 (corresponding to the number of frames in the video turned into time series).

%%%%%%%%%%%%%%%%%%%%%%%Start Time Series Classification%%%%%%%%%%%%%%%%%%%%%%%%%%% 
 
%    \noindent\textbf{Time Series Classification.}
%    We select the multivariate time series classification methods based on their performance on the multivariate TSC archive \citep{DBLP:journals/datamine/RuizFLMB21}. As described in the previous section, we employ ROCKET and MINIROCKET \citep{Dempster2020, Dempster2021MINIROCKETAV} as well as the Fully Convolutional Network (FCN) and Residual Network (ResNet)  \citep{Fawaz2019}. 

%%%%%%%%%%%%%%%%%%%%%%%End Time Series Classification%%%%%%%%%%%%%%%%%%%%%%%%%%% 

\section{Data Mining Challenges and Solutions in the Context of BodyMTS}
\label{sec:dmc}
    In this section we present the data mining challenges posed in the context of this application's requirements. We discuss the challenges and solutions for each stage of BodyMTS as shown in the Figure \ref{fig:overview}. 
    
    \begin{itemize}
    
     \item \textbf{Data size.} Video data is large and requires extensive memory and storage resources as well as high end computation machines. 
     %Deep learning methods which directly work on video data require expensive computation resources such as GPU and memory for training and prediction. \\
     
    \textit{Solution:} We investigate approaches to reduce the video size, such as frame cropping and centering, increasing the video compression ratio (using CRF) and reducing the video to time series. We find that there are settings that give a significant data reduction (70\%), and still preserve classification accuracy. Experiments investigating this challenge are presented in Section \ref{sec:noiseexp}.
    
    \item \textbf{Noisy data.} The steps of video data capture and reduction, pose estimation or time series pre-processing can reduce the quality of the data by introducing noise. BodyMTS as well as deep learning methods that work directly with video are affected by the level of data noise (e.g., blurred, poor quality videos). The accuracy of OpenPose is also directly impacted by video quality. \\
   %Poor quality videos affect the confidence of pose estimation which in turn affects the classifier performance. \\
    \textit{Solution:} We examine the impact caused by data reduction on accuracy. We evaluate different settings for training and predicting on high quality data, as well as training on high quality, and predicting on lower quality data (more noise) to simulate realistic use cases. We also evaluate different settings for pose estimation and time series processing. We find that there are data and model settings that are robust to noise, preserve classification accuracy and have near real-time prediction. Experiments investigating this challenge are presented in Section \ref{sec:noiseexp}.
    
    \item \textbf{High dimensionality of multivariate time series and scalability of existing classifiers.}
    Pose estimation libraries such as OpenPose track multiple key-points on a human body (25 for OpenPose), which may lead to large scale multi-dimensional time series data. \\
    \textit{Solution:} We overcome this challenge by consulting domain experts as well as evaluating recent approaches to automatically select useful channels. Out of 25 body parts, we show that involving only 8 upper body parts achieves an accuracy of 87\% on average for Military Press. We also explored techniques such as skipping frames in the input video during pose estimation.
    We examine recent scalable multivariate time series classifiers and find that ROCKET behaves the best among state-of-the-art methods, with regard to both accuracy and training/prediction time. Experiments investigating this challenge are presented in Section \ref{sec:tscexp}.
    
    %Additionally, due to the easy availability of smartphones and social media platforms there has been a significant increase in the amount of video data.
    
    %In the previous experiments, we found that even after tuning the parameters mainly responsible for enhancing the confidence of OpenPose do not necessarily result in improving the classifier performance. 
    
    \item \textbf{Segmentation of time series.} The input video is a sequence of 10 repetitions of one exercise, which simulates actual use cases. The classification is done on each individual repetition, thus a segmentation step is required to break down the data into single reps.
    The current method utilizes peak detection on pose estimation time series to obtain data for each repetition. This approach is prone to noisy fluctuations (due to pose estimation errors) and may lead to incomplete repetitions. Additionally, it is not known in advance which body parts to utilize for segmentation. \\
    \textit{Solution:} We investigate different channel selection methods to capture the subset of body parts that is more relevant for the target exercise. We also analyse simple segmentation techniques directly on video, versus segmentation based on pose estimation time series data. We find that directly splitting the video into equal parts works reasonably well, although it is less accurate than using peak detection on pose estimation time series. Experiments investigating this challenge are presented in Section \ref{sec:segcompare}.

    %\item \textbf{Scalability of multivariate time series classification.}
    %The data obtained is sequential and high dimensional in nature. It becomes crucial to investigate the latest and accurate MTSC algorithms with minimum train/test time. We currently explored techniques such as skipping frames in the video signal and channel selection to reduce the data size while maintaining the same accuracy. \\
    %\textit{Solution:}
    
    \item \textbf{Data privacy and security.}
    Capturing video data comes with implicit challenges such as privacy and data security. It becomes absolutely critical to design approaches that  maintain the individual anonymity.\\
    \textit{Solution:} We turn videos into pose time series, which removes any visual clues about the identity of users. Pose estimation could in principle be used for height estimation, but this reveals little about the identity of users.

\end{itemize}

    \section{Experiments}
    \label{sec:exp}
    
    This section presents an empirical evaluation of our method and is organized around the data mining challenges  discussed in the previous section.
    In Section \ref{sec:dlcompare} we compare the performance of different deep learning methods for video classification versus BodyMTS with regard to accuracy and computational efficiency. %We select state-of-the-art classifiers and present their accuracy on our dataset. 
    %We first evaluate several deep learning methods and  compare their performance with BodyMTS in terms of total time, accuracy and required resources. 
    We also evaluate the impact of two segmentation techniques on the performance of BodyMTS and the best deep learning method. We further compare the impact of video quality on the best deep learning method versus BodyMTS.
    Section \ref{sec:noiseexp} addresses challenges raised by noisy data and data size. We analyze the robustness of BodyMTS against different sources of noise which can be broadly grouped into 3 categories: (1) video data capture; (2) OpenPose parameters; (3) time series data pre-processing. 
    %We generate noisy data with different levels of noise and analyze the impact on accuracy. 
    %We change the video quality through video attributes such as CRF and resolution. We further evaluate the impact of tuning OpenPose parameters on the BodyMTS accuracy. We show the impact of changing classifier parameters and data pre-processing in the Appendix.
    %We compare the original size of the video with the multivariate time series data obtained after the pose estimation. We also address how to further reduce the video data size by introducing noise (degrading the video quality) without a major drop in the accuracy. 
    Section \ref{sec:tscexp} addresses high dimensionality and scalability for time series classification. We present the accuracy and compare the total execution time for different classifiers. We further evaluate the impact of utilizing different subsets of body parts on BodyMTS accuracy.
    
    We have not included any experiments to address the issue of data privacy and security as BodyMTS works directly on the multivariate time series data and hence reasonably safeguards the identity of the user.
    
    \subsection{BodyMTS versus Direct Human Action Recognition Classifiers}
    \label{sec:dlcompare}
    In this section, we compare BodyMTS with state-of-the-art methods for human activity recognition from videos. These methods employ deep learning architectures and have shown good performance on several benchmark datasets such as UCF101, Kinetics-400 and Kinetics-600. We selected a few methods based on their performance, execution time and resources required. The following section provides a brief overview of the selected methods.

    \begin{itemize}
    \item \textbf{C2D} \citep{fan2020pyslowfast} stands for 2D convolutional based model. All convolutions are performed on each of the frame. ResNet-50 and ResNet-101 can be used as the backbone architectures for C2D.
    
    \item \textbf{I3D} \citep{Carreira2017} stands for inflated 3D Convolutional network. They work by inflating the kernels of C2D models in order to capture the temporal information. These models are computationally expensive due to an increased number of computations.
    
    \item \textbf{SlowFast} \citep{Feichtenhofer2019} architecture uses two pathways to perform activity recognition from videos. The slow pathway captures the spatial information at low frame rates and the fast pathway captures the temporal information at high frame rates. The fast pathway requires less computation because it uses a backbone network with reduced channel sizes, which is normally 8 times smaller than slow path backbones. The information from both the pathways is fused by lateral connection. The backbone architecture of SlowFast can be a 3D ResNet-50, 3D ResNet-101, a Non-local Network or a combination of these. 
    
    \item \textbf{Non-local Network} \citep{Xiaolong2017} is  used to capture the long range dependencies by enhancing the large receptive fields. They can be integrated as a generic building block with most deep learning architectures. In the experiments by \citep{Feichtenhofer2019} they have been combined with existing standard backbone models such as ResNet-50 or ResNet-101.
    
    \item \textbf{X3D} \citep{Feichtenhofer2020} progressively expands a 2D CNN along multiple axis in space, time, width and depth. It uses the progressive forward expansion followed by backward contraction. The axis is selected based on the performance of the model.
    \end{itemize}
    
    \subsubsection{Deep Model Architectures}
    We use the SlowFast library \citep{fan2020pyslowfast} to evaluate the above mentioned models. Table \ref{table:model_config} shows the number of frames, sampling rate and the frame cropping size for these architectures. The column ``Model Config" refers to the name of the config file in the SlowFast repo\footnote{\url{https://github.com/facebookresearch/SlowFast}}. All models are initialized with the weights pre-trained on Kinetics-400. In column ``Model config'', \textit{R50} indicates the ResNet-50 has been used as the backbone architecture; the numeric values in the format $X\times Y$ where $X$ indicates the number of frames and $Y$ indicates the sampling rate, e.g. C2D\_8$\times$8\_R50 is a C2D model which utilizes a total of 8 frames with sampling rate of 8 from a video clip.
    All the experiments are executed on an Ubuntu machine with a single GPU (NVIDIA TITAN XP 12 GB, Ubuntu 18.04.5 LTS, AMD Ryzen 7 1700X Eight-Core Processor). 
    We evaluate different values of batch size to utilize the maximum amount of GPU. After multiple iterations we found a batch size of 5 suitable for all the architectures to avoid getting out of memory errors.
    % We use default model configurations to build DNN models.During training, 2 snippets are randomly sampled from videos. 
    We use 10\% of training data as validation set. The validation set is chosen such that participant ids of training and validation are not overlapping. 
    %We report the accuracy over a single split of data. 
    Table \ref{table:model_config} shows the selected models: SlowFast, C2D, X3D, and I3D.
    In the case of testing, 10 frames are uniformly sampled from each clip consisting of a single repetition along the temporal axis. Each frame  underwent top/left, middle and bottom/right cropping thus giving a total of 30 frames for each video. The final softmax score is averaged over the 30 frames to give the final prediction. We report the training and testing time for each of these architectures. Please refer to \citep{Feichtenhofer2019} for more details on the data pre-processing and configurations.
    Note that directly running these models with the default parameters given in the SlowFast repository does not lead to good results. A considerable amount of engineering effort was spent in tuning the hyperparameters: learning rate, batch size, epochs, warm-up epochs and weight decay. All other hyperparameters remain unchanged. All the models are trained with Stochastic Gradient Descent (SGD) with momentum 0.9 for 10 epochs. 
    %In the next section, we discuss the average accuracy of these models on three train/test splits.
    
    \begin{table}[h!]
    \centering
    \begin{tabular}{lllll}
    \toprule
    Model Name    & Model config                      & No. of Frames & Sampling Rate & Size\\ \midrule
    C2D           &  C2D\_8$\times$8\_R50             & 8   & 8 & $224^2$   \\
    I3D           &  I3D\_8$\times$8\_R50             & 8   & 8 & $256^2$   \\
    SlowFast      & SLOWFAST\_4$\times$16\_R50        & 32  & 2 & $256^2$   \\
    SlowFast + NL & SLOWFAST\_NLN\_4$\times$16\_R50   & 32  & 2 & $256^2$   \\
    X3D-M         & X3D\_M                            & 16  & 5 & $256^2$   \\
    % R(2+1)D       & R2PLUS1D\_16$\times$4\_R50        & 16  & 4 & $224^2$   \\
    \bottomrule
    \end{tabular}
    \caption{Selected Deep Learning Activity Recognition Models and their configurations.}
    \label{table:model_config}
    \end{table}
    
    \begin{table}[h!]
    \centering
        \resizebox{\columnwidth}{!}{%
        \begin{tabular}{lccc}
        \toprule
            Model                    & Accuracy     & Total Test time (mins)/ & Training time (mins) \\
            & & Test time per clip of 10 reps (mins) &   \\ \midrule
            C2D                      & 0.67 $(\pm0.021)$         & 23/0.38               & 19 \\
            I3D                      & 0.79 $(\pm0.036)$         & 32/0.53               & 23 \\
            SlowFast                 & 0.83 $(\pm0.021)$         & 29/0.48               & 52 \\
            SlowFast + NL            & 0.83 $(\pm0.020)$         & 27/0.45               & 59 \\
            X3D-M                    & 0.78 $(\pm0.012)$         & 41/0.68               & 126 \\
            \midrule
            BodyMTS (frame-step=1)                 & 0.87 $(\pm0.026)$        & 22/0.36               & 52 \\
            BodyMTS (frame-step=3)                  & \textbf{0.85} $(\pm0.029)$         & \textbf{12/0.20}               & \textbf{26} \\
            \bottomrule
        \end{tabular}
        }
        \caption{Average accuracy, total testing time, time per testing clip and total training time for different architectures over 3 train/test splits. The average duration of all clips in training and testing is 65 mins and 30 mins respectively. Note: all deep learning models are pre-trained on Kinetics-400. For accuracy without pre-training, see Table \ref{table:dl_bd_comp}.}.
        \label{table:dl_accuracy}
    \end{table}

    \subsubsection{Results of BodyMTS vs Deep Learning Models}
    
    Table \ref{table:dl_accuracy} reports the average accuracy and average running time of the deep learning models evaluated. Note that this accuracy was obtained with significant model engineering effort and with using pre-trained weights from the Kinetics-400 benchmark for all the deep models. 
    As seen from Table \ref{table:dl_accuracy}, C2D performed the worst whereas SlowFast and SlowFast + NL achieve the highest accuracy, followed by I3D and X3D-M. We select the best architecture model which is SlowFast in this case and compare its performance with BodyMTS. 
    We use the short name SlowFast for this model in the subsequent sections. %Table \ref{table:dl_bd_comp} shows the comparison of SlowFast with BodyMTS based on accuracy, execution time, cost and storage space. 
    In the test data there are a total of 60 clips each with 10 repetitions on average. These clips were fed one by one to the pre-trained model. The testing time for each model is the summation of total time taken for data pre-processing (which includes segmentation), model loading and classification. Table \ref{table:dl_accuracy} shows the total testing time taken for 60 clips for each model. We also reported the average testing time over single clips of 10 repetitions. The training time shown is the summation of time taken for data pre-processing and model training for all the clips in the training data. 
    
    \noindent\textbf{Accuracy.} Table \ref{table:dl_bd_comp}  reports the average accuracy over three splits for SlowFast (0.83) and BodyMTS (0.87). BodyMTS achieves higher accuracy  with minimal model engineering effort. We note that the default architecture of SlowFast is meant for classifying large scale video datasets such as UCF-101 or Kinetics-600 and so there is a higher chance of overfitting. This is substantiated by observing that there is a very significant drop in the accuracy of SlowFast when removing the pre-trained weights as shown in Table \ref{table:dl_bd_comp}, with the accuracy dropping by 60 percentage points, from 0.83 to 0.25. It is only after using the pre-trained weights on Kinetics-400 and other model engineering steps that SlowFast reaches an average accuracy of 0.83. 
    
    \noindent\textbf{Execution Time.} We report the total training and testing time for both the models in Tables \ref{table:dl_accuracy} - \ref{table:dl_bd_comp}.
    The total duration of all the videos (both training and test) is 95 minutes.
    We observe that the combined train/test time of BodyMTS (OpenPose + data pre-processing + training/testing) is around 74 minutes whereas the combined train/test time of SlowFast is around 86 minutes. Prediction time is 22 minutes for BodyMTS and 27 minutes for SlowFast. This shows that BodyMTS is faster than SlowFast for both training and prediction. Additionally, using a frame step of 3, the combined train/test time of BodyMTS goes down to 38 minutes which is significantly faster than the time taken for SlowFast. Excluding the execution time of OpenPose, BodyMTS only takes a total of 2 minutes including both the training and testing time for the classifier. Additionally, for a single clip consisting of 10 repetitions BodyMTS takes a total time of 12 seconds, whereas SlowFast takes 29 seconds on average. Hence BodyMTS is overall faster than SlowFast and can deliver near real-time predictions. %Reducing the frame-rate in OpenPose can lead to a much reduced training and testing time for BodyMTS. The total training time comes down to 37 mins and 26 mins for frame-step 2 and 3 respectively whereas the total testing time comes down to 17 mins and 12 mins for frame-step 2 and 3 respectively. 
    %Therefore using a frame-step of 3, can reduce the training time by a factor of 2 and by a factor of 5 in testing. 
    %BodyMTS is faster during deployment as OpenPose can process frames at a speed of 40 fps (for a normal video at 30fps) making it plausible to provide near real time feedback to the user.
    % Training time consists of the data pre-processing time and the model training time for both the cases. Data pre-processing in case of BodyMTS consists of running the OpenPose followed by segmentation and saving the data and feeding it to the training data. Data pre-processing in case of the DL methods consists of the randomly clipping the files 
    
    % \iffalse
    % \begin{table}[h!]
    % \centering
    % \begin{tabular}{lcccc}
    % \toprule
    % Model              & Inference time for single video (sec)  & Total & hit@1 \\ \midrule
    % C2D                & 0.40                           &                       & 77.04\\
    % I3D                & 0.58                           && 85.02\\
    % SlowFast           & 1.69                           && 84.86\\
    % SlowFast + NL      & 0.89                           && 86.02\\
    % X3D-M              & 1.19                           & & 85.86\\
    % R(2+1)D            & 1.25                           && 83.53\\
    % \midrule
    % BodyMTS            & 2.16 \\
    % \bottomrule
    % \end{tabular}
    % \caption{Inference processing time per video. Classification time per video is estimated using batch size of 30.}
    % \end{table}
    % \fi
    
    \begin{table}[h!]
    \centering
        \resizebox{\columnwidth}{!}{%
        \begin{tabular}{p{4cm}|p{5cm}|p{4cm}}
        \textbf{Step} & \textbf{SlowFast} & \textbf{BodyMTS (frame-step=3)} \\
             \toprule
            Data Size                            & 213 MB (videos)             & 28 MB (time series) \\  \midrule
            % OpenPose                             & 1 Hour 14 mins             & NA \\
            % Data Pre-processing Time                 & NA                          & 74 mins(frame-step 1), \newline 54 mins(frame-step 2), \newline 38 mins(frame-step 3) \\
            Training Time                            & 59 mins                     & 26 mins \\
            Testing Time                             & 27 mins                     & 12 mins \\\hline
            \textbf{Average accuracy over three splits}                             & \textbf{0.83 (with pre-training on Kinetics-400, epochs=10)}\newline\newline \textbf{0.25 (no pre-training, epochs=50) }                      & \textbf{0.85} \newline  \\\hline
            % \textbf{Average accuracy \newline over 3 splits}                             &                      0.83 & \textbf{0.87}  \\ \hline
            Infrastructure                       & GPU                         & GPU or CPU \\ \bottomrule
        \end{tabular}
        }
        \caption{Comparison of SlowFast and BodyMTS approaches in terms of time taken and resources required. The total duration of all clips in training and testing is 65 mins and 30 mins respectively.}
        \label{table:dl_bd_comp}
    \end{table}
    
    \noindent\textbf{Cost.} All deep learning methods require the use of GPU. In our case, a single GPU was utilized for the deep learning models to compare their performance. Increasing the number of GPUs will lead to a reduced execution time but at the cost of expensive infrastructure.  However, running deep learning models without any GPU will lead to significant increase in  training/testing time. The CPU version of OpenPose takes roughly about 15sec/frame \citep{openpose2019}, however, the recent lightweight implementations of OpenPose makes it possible to reach real-time inference on CPU with negligible accuracy drop \citep{osokin2018lightweight_openpose}. We recently found that libraries such as OpenVINO \footnote{\url{https://docs.openvino.ai/2019_R1/_human_pose_estimation_0001_description_human_pose_estimation_0001.html}} makes it possible to execute OpenPose on the CPU machines and TensorFlow Lite\footnote{\url{https://www.tensorflow.org/lite/examples/pose_estimation/overview\#performance_benchmarks}} supports running pose estimation models directly on the mobile phone in real-time.
    % We recently found that the deep learning library TensorFlow Lite \footnote{\url{https://www.tensorflow.org/lite/examples/pose_estimation/overview#performance_benchmarks}} supports running pose estimation models directly on the mobile phone in real-time. 
    This is interesting for future work, but a different workflow than the one we use here. Therefore, it is possible to reduce the computation footprint of BodyMTS more than the corresponding deep learning models for video-based exercise classification.
    
    \noindent\textbf{Storage Space.} 
    Table \ref{table:dl_bd_comp} shows the initial data size for SlowFast and BodyMTS. Because of the large size of videos there is high cost involved to store the videos. However, videos do not need to be stored for BodyMTS, once the time series are extracted. Since the data is just a sequence of numbers (time series), there can be large savings in storing this data. 
    %Moreover, in case of resource-constrained devices such as mobile phones, it is not possible to store large video datasets. 
    Even when the data is sent to the cloud server, there can be large savings in terms of bandwidth required to transfer the data. 
    %Lastly, videos need to be compressed before transfer to the cloud which can lead to loss of information.
    
    % GI Add engineering effort
    \noindent\textbf{Practical Aspects.}
    From Table \ref{table:dl_accuracy} and Table \ref{table:dl_bd_comp}, we see that BodyMTS has higher accuracy than the deep learning methods. Deep models may suffer from high training and test time depending upon the data size as well as require high engineering effort to tune hyper parameters such as the number of epochs, learning rate, etc.
    Increasing the number of GPUs may lead to decrease in execution time for the deep learning models, however, this may not be always possible due to cost. In contrast, due to the lightweight nature of the time series classifier, BodyMTS does not require GPU resources (using lightweight OpenPose) and can be trained/tested within a fraction of the runtime of SlowFast, on a single CPU machine \citep{dempster2019rocket, osokin2018lightweight_openpose}. 
    %Moreover, running deep learning based methods on resource-constrained devices such as mobile phones may be impractical due to computation resources (e.g. GPU, memory) and large data size. Additionally, in case of BodyMTS, there are huge savings possible in terms of storage space due to the low storage requirements of time series data. Therefore, BodyMTS seems very promising for video-based exercise classification. We note that depending upon the type of physical exercise and the requirements of the application, further investigation may be needed to select the best approach.
    
    \subsubsection{Impact of Segmentation}
    \label{sec:segcompare}
    In this section we analyze the impact of two segmentation techniques on the accuracy of BodyMTS and SlowFast. As stated earlier in the data pre-processing step, segmentation of the video data is required to obtain the individual sample for each repetition for train/test data. We consider two scenarios here:
    \begin{itemize}
        \item When the start and the end time of each repetition is known in advance. This information can be utilized to easily segment the video data into individual repetitions. To get this information we use pose estimation and peak detection techniques.
        \item When the number of repetitions are known in advance. Dividing the total duration of the video clip by the total repetitions can approximately segment the individual repetition assuming that the participant takes consistent time to complete each repetition. This approach does not require pose estimation information.
        % Moreover, in case of video figuring out the number of repetitions beforehand number will not be difficult in case of video.
    \end{itemize}
    
    \begin{table}[h!]
    \centering
        \resizebox{0.9\columnwidth}{!}{%
        \begin{tabular}{lcc}
        \toprule
        \textbf{Segmentation Approach} & \textbf{SlowFast Accuracy} & \textbf{BodyMTS Accuracy} \\ \midrule
            With Pose Estimation                              & 0.83             & 0.87 \\
            Without Pose Estimation                  
                    & 0.77             & 0.80 \\
            \bottomrule
        \end{tabular}
        }
        \caption{Average accuracy obtained by SlowFast and BodyMTS for two different segmentation approaches for three train/test splits. }
        \label{table:seg_accuracy}
    \end{table}
    
    Table \ref{table:seg_accuracy} shows the average accuracy of SlowFast and BodyMTS using two different segmentation approaches over three train/test splits. BodyMTS achieves  4 percentage points higher accuracy than SlowFast when the data from pose estimation is utilized for segmentation. When the segmentation is done by equally dividing the total duration by the total number of repetitions, BodyMTS still achieves  3 percentage points higher accuracy than SlowFast. Therefore, from the above results accuracy obtained is higher when the repetitions are correctly segmented using pose estimation than the scenario where the segmentation is performed by equally dividing the total duration with the total number of repetitions.  This suggests that repetitions may not be fully captured as this is just an approximate way to obtain each repetition. The idle time at the start or the end of the video clip as well as the variation in the duration of each repetition can affect the segmentation. %In the future, we plan to investigate further techniques for better  segmentation of the video data.
    
    \subsubsection{Impact of Video Quality Noise}
   We now analyze the impact of video quality noise on the deep learning models, as well as BodyMTS. 
   %We select SlowFast as it is the most accurate as shown in Table \ref{table:dl_accuracy}. 
   We do this by changing the CRF video property as discussed in detail in Section \ref{sec:noiseexp}. Higher value of CRF leads to a drop in the quality of the video and vice versa. We compare the accuracy at different values of CRF: the default is set at 23 and we test a higher quality of video at CRF 16 and degrading the quality of the video all the way to CRF 34. 
   %As shown in Table \ref{table:crf_accuracy} a CRF of 28 provides a good threshold level until which the video quality can be decreased without significantly affecting the accuracy of BodyMTS.
    \begin{table}[h!]
    \centering
        \resizebox{0.8\columnwidth}{!}{%
        \begin{tabular}{lcc}
        \toprule
        \textbf{CRF} & \textbf{SlowFast Accuracy} & \textbf{BodyMTS Accuracy} \\ \midrule
            16                              & 0.84             & 0.87 \\
            23 (default)                    & 0.83             & 0.87 \\
            28                              & 0.82             & 0.85 \\
            34                              & 0.81             & 0.81 \\
            % 40                              & 0.84             & 0.71 \\

            \bottomrule
        \end{tabular}
        }
        \caption{Average accuracy obtained by SlowFast and BodyMTS for varying video quality (CRF from 16 to 34) over 3 train/test splits.} %The total video size is reduced from 213 MB at CRF 23 to 76 MB at CRF 28.}
        \label{table:noise_acc_comp}
    \end{table}
    Table \ref{table:noise_acc_comp} shows the average accuracy obtained with SlowFast and BodyMTS at different CRF over three train/test splits. 
    We observe that at the default value of CRF 23, BodyMTS achieves higher accuracy than SlowFast by 4 percentage points. Reducing the video quality by increasing the CRF affects both methods, with the accuracy decreasing but still  above 80\% which is desirable as described in Section \ref{sec:app_require}.
    %At CRF 28, there is a drop of 2 percentage points in BodyMTS accuracy and 1 percentage point in SlowFast accuracy. However, BodyMTS still achieves better accuracy than SlowFast. 
    % The above results suggest that BodyMTS can still perform better or match the performance of SlowFast up to a threshold of CRF 30. 
    % At CRF 40, (see Figure \ref{fig:goodvsbad}), there is an improvement in the accuracy of SlowFast as compared to the accuracy at CRF 23. Therefore, it would be interesting to explore what part of the image model is looking at using the saliency maps. Using the initial analysis in our case we observed that SlowFast is looking at the background which makes no sense.  
    % This suggests that BodyMTS is more robust than SlowFast to noise arising from varying video quality. 
    For future work, it would be interesting to study whether video quality metrics such as VMAF \cite{netflixvmaf} could also be used to identify an application specific threshold beyond which the video quality is too poor for inclusion in this task. In the Appendix, we investigate a few video quality metrics and the corresponding BodyMTS accuracy.
    
    \textbf{Takeaway:} 
    The previous experiments suggest that BodyMTS is more accurate, significantly faster and more cost-efficient when compared to the best deep learning method, SlowFast. 
    %BodyMTS is also robust to lower quality video.
    % VMAF seems promising as a video quality metric and correlates well with the final time series classifier accuracy.

    \subsection{Robustness Analysis: Impact of Noise on BodyMTS}
    \label{sec:noiseexp}

    In this section, we analyze the robustness of BodyMTS against different sources of noise that may occur in this application. %Our main motivation is to evaluate and analyze the impact of most probable noise types on the accuracy of BodyMTS. 
    %Since it is impossible to capture such data with all kinds of realistic noise, we simulate this by introducing the noise into the original data. 
    % We generate data with different levels of noise and analyze its impact on the classifier accuracy. 
    These sources of noise can be broadly classified into 3 categories: (1) video data capture; (2) OpenPose parameters; (3) time series data pre-processing.
    %We use ROCKET as the multivariate time series classifier in BodyMTS, since it is accurate, fast and more scalable than other classifiers \citep{Dempster2020}. Also, ROCKET does not require heavy computing  resources and  can  be  easily  trained  and tested  on  a  standalone CPU-based machine. 
    While studying the impact of noise we address the following questions:

    \begin{enumerate}
    
    \item How does the noise from different sources such as video capture quality, OpenPose estimation and data pre-processing affect the classifier accuracy? 
    
    \item Is it possible to reduce the quality of videos but still keep the same accuracy? Are there possible benefits in terms of saving storage space by reducing the data size? 
    %\item Is it possible to improve the accuracy of time series classifiers (TSC) in spite of noise present in the data? Does having an awareness of the noise level (as estimated by OpenPose confidence) help us in designing more robust TSC?
    
    \end{enumerate}
    
    We address these questions by generating different types of noisy videos with varying levels of noise. We degrade or enhance the quality of videos by changing the resolution and bit rate. Also, we analyze the impact of OpenPose parameters that are most responsible for affecting the quality of estimation. In addition, we explore parameters of OpenPose that can be tuned to reduce the overall execution time of BodyMTS. We further analyze how much the quality of videos can be reduced by changing the compression level without sacrificing accuracy. 
    %This can possibly help in running the application on resource constrained devices such as mobile phones where there is less storage space. 
    %We quantify the amount of noise by using standard video quality metrics.
    We consider two scenarios: 1. Adding noise to both training and testing data. 2. Adding noise to testing data and keeping the training data intact.
    %The rationale behind this choice is that if the current data is noisy, the data in the future may also be noisy.
    %2) When the noise is added to testing data only. The helps us in analyzing how the model trained on clean data behaves when presented with noisy data.
    %The next sections provide an overview of different types of noise we have studied.

%%%%%%%%%%%%%%%%%%%%%%%%%%%%%%%%%%%%%%%%%%%%%%%%%%%%%%%%%%%%%%%%%%%%%%%%
%%%%%%%%%%%%%%%%%%%%Video quality%%%%%%%%%%%%%%%%%%%%%%%%%%%%%

    \subsubsection{Data-Capture Noise}
    \label{sec:datacapturenoise}
    In this section, we study the impact of noise coming from data capture. This can be further categorized into two types: video quality and the recording conditions. We describe each of them in detail below.\\

    \noindent\textbf{Video Quality.}
    The motivation behind studying the video quality is that videos captured in the wild can range from very poor quality to high definition quality. %Here we study the impact of noise coming from the quality of video. %The objective is to analyze the impact of varying levels of video quality on BodyMTS.
    For modern smartphones the camera quality is much more efficient than it used to be 10 years ago. Nevertheless, recorded videos can still have a low quality because of the compression required to send the data to the cloud service where further processing takes place. 
    We generate varying quality of video data with different levels of noise by tweaking the video properties such as the bit-rate and resolution. %Both of these are responsible for determining the quality of video. 
    % In addition, there are other properties such as codecs and containers which may influence the quality, however we do not consider them in the current analysis. 
    
    % The goal behind generating these videos by changing video properties is to replicate the real-world behavior where the videos can range from poor quality recorded with low-end phones to high quality recorded with HD/DSLR cameras.
    
    % The major source of noise in the video can be attributed to various properties 
    % such as the bit-rate, resolution, encoding methods, frame rate, etc. We only take into account those video properties that determine the quality of compressed videos.  
    We use FFmpeg \cite{ffmpeg} to obtain noisy videos by modifying the above properties. This is an open-source, widely used tool used for manipulating and modifying videos. 
    % Since it may not be entirely feasible to send the high quality videos as that will utilize more bandwidth and require high computation. 
    Additionally, OpenPose estimation confidence is directly proportional to the quality of the video. Lower quality videos lead to low confidence of the body-parts location estimation, which ultimately may affect the classifier accuracy. %We analyze this aspect in Section \ref{sec:noiseexp}. 
    
    \begin{itemize}

    \item \textbf{Resolution} refers to the number of pixels in each image.
    The higher the resolution, the more pixels and hence the better video quality. In our experiments, we downscale the original resolution of the videos to measure the impact on classifier accuracy. Note that the original videos of the Military Press were recorded at HD quality with 720p resolution.
    
    \item \textbf{Bit-rate} refers to the amount of information processed per second to represent a video. Higher bit-rate means more information, which means better quality, but also a higher video file size. We use the constant rate factor (CRF) \cite{ffmpeg}, which is a rate control mode, to change the number of bits per frame. We examine the impact of bit rate by changing the CRF property of videos. Higher CRF leads to lower quality videos.
    % \item \textbf{Color space} defines the color display of video when watched on a display device. We change the color space of the videos from the original RGB format to gray-scale to analyze its impact on the classifier accuracy.
    
    % \item \textbf{Frame-rate} is defined as the number of frames processed per second by the media player and determines the fluidity of the video. The common accepted frame-rate is 30fps, which renders a smooth flow.
    
    % \item \textbf{Codecs} are used to compress and decompress video data. Different codecs come with different compression algorithms. The compression is generally lossy which leads to saving of storage space however at the cost of losing information. Common codecs include H.264 and HEVC. H.264 uses lossy compression techniques and is the most common video codec. 
    % \textbf{Containers} are used to hold the contents of video along with metadata. They have extensions like .mov, .avi, .mkv or .mp4 etc.
    \end{itemize}
    In addition to the above properties, we also alter the color space and frame rate to analyze their impact on the BodyMTS accuracy.\\
    % We further quantify the quality of videos by using standard video quality metrics such as VMAF \citep{netflixvmaf}, PSNR and SSIM. We particularly focused on the VMAF score as it takes into account the user perceptibility. Detailed results for these metrics are shown in the Appendix. \\

%%%%%%%%%%%%%%%%%%%%%%%%%%%%%%%%%%%%%%%%%%%%%%%%%%%%%%%%%%%%%%%%%%%%%%%%
%%%%%%%%%%%%%%%%%%%%%%Recording conditions%%%%%%%%%%%%%%%%%%%%%%%%%%

    \noindent \textbf{Recording Conditions.}
    Several factors during the video recording can affect the way a video is recorded which ultimately can affect the confidence of OpenPose. We categorize common recording conditions into different types and provide a brief description below.
    
    \begin{itemize}
    
    \item \textbf{Camera Settings}: This may include factors such as orientation, viewing angle, zoom of the camera, distance between the participant and the camera or whether the participant is centralized or not, etc. 

    \item \textbf{Background}: Depending upon the conditions such as whether the participant is executing the exercise in an indoor or outdoor settings, this may influence the quality of video. For instance, in a gym setting factors like multiple people, background noise (pictures having humans), clothing (background color same as clothing) and lighting may affect the final output of videos.
    %\item \textbf{Other}: Other factors like time and location of the day can also have substantial impact on the data quality.
    \end{itemize}
    
    Apart from the above, there may be other unaccounted factors which can influence the recording conditions. %Nonetheless, it is to be noted that testing out all scenarios is not feasible. 
    In our data, the distance to the camera and the background can change, and these variations did not affect the accuracy of BodyMTS. Nevertheless, we note that  
    BodyMTS expects certain conditions to be met (e.g., regarding camera being stable) as listed out in Section \ref{sec:app_require} before deploying.
%    We discuss some of these conditions in our experiments on noise related to video capture.    
%%%%%%%%%%%%%%%%%%%%%%%%%%%%%%%%%%%%%%%%%%%%%%%%%%%%%%%%%%%%%%%%%%%%%%%%
%%%%%%%%%%%%%%%%%%%%%%%%%End%%%%%%%%%%%%%%%%%%%%%%%%%%%%%%%%%%%%%%%%%%%

%%%%%%%%%%%%%%%%%%%%%%%%%%%%%%%%%%%%%%%%%%%%%%%%%%%%%%%%%%%%%%%%%%%%%%%%
%%%%%%%%%%%%%%%%%%%%OpenPose parameters%%%%%%%%%%%%%%%%%%%%%%

    \subsubsection{OpenPose Parameters}
    
    BodyMTS utilizes OpenPose to obtain the coordinates information for major key-points in a human body. OpenPose is a full-fledged, 2D pose estimation system that includes many parameters that affect accuracy, optimization, display and output format. Here the objective is to evaluate and tune the parameters which may influence the accuracy and efficiency of OpenPose which ultimately affects the running time and accuracy of BodyMTS. %We examine the impact of changing these parameters on the BodyMTS accuracy. 
    A short description of these parameters is provided below:
    
    \begin{itemize}
    \item \textbf{Frame-step}: an integer value indicating the number of frames to skip during the estimation.
    
    % \item \textbf{Number-people-max}: this parameter limits the maximum number of people detected by keeping the people with top scores. The score is based on person area over the image, body part score, as well as joint score (between each pair of connected body parts)
    
    \item \textbf{Net-resolution}: increasing this may  increase the accuracy while also increasing the execution time.
    
    \item \textbf{Scale-number}: this parameter indicates the number of scales to average.
    
    % \item \textbf{Scale-gap}: this parameter indicates the scale gap between scales.
    
    \end{itemize}

    The parameters \emph{net-resolution, scale-number} are directly responsible for the accuracy of OpenPose.
    % whereas \emph{number-people-max, frame-step} can be used to speed up the execution of OpenPose. The parameter \emph{number-people-max} can be further used to remove the extra persons in the background based on the confidence score. 
    OpenPose fails to detect the body parts in case the person is not facing the camera or the body part is cropped accidentally or the camera fails to capture the whole body. In the Military Press execution, the participant is not facing the camera and so OpenPose fails to detect the coordinates for eyes and nose. Nonetheless, these body parts are not involved in the physical movement and so have no impact on the accuracy.

%%%%%%%%%%%%%%%%%%%%%%%%%%%%%%%%%%%%%%%%%%%%%%%%%%%%%%%%%%%%%%%%%%%%%%%%
%%%%%%%%%%%%%%%%%%%%%%%%End%%%%%%%%%%%%%%%%%%%%%%%%%%%%%%%%%
    
    All the experiments are performed using the sktime \citep{sktime} version of ROCKET on an Ubuntu 18.04 system (16GB RAM, Intel i7-4790 CPU @ 3.60GHz). %We use the same pre-processing steps as discussed in  Section \ref{sec:proposed_approach}. 
    %We report the average accuracy over three splits of data to evaluate the impact of noise. We start by generating varying levels of video quality and studying the impact on  accuracy. In the subsequent section, we examine the impact of OpenPose parameters on the BodyMTS accuracy. 
    
   \subsubsection{Results for the Impact of Video Quality on BodyMTS}
    In this section we analyze the impact of changing the video quality on the BodyMTS accuracy. We start by changing the bit-rate and resolution both of which are responsible for determining the quality of video. 
    We further present the total size of the video data obtained after altering each property. We note that in the following experiments we consider the case when both the train and the test data have been impacted by the changes in video quality. \\
    
    \noindent\textbf{Reducing the resolution}: We reduce the original resolution in steps of one-half, one-third of the original resolution and evaluate its impact on the classifier accuracy. Table \ref{table:reso_accuracy} shows the impact of various resolutions on the classifier accuracy and data size.

    \begin{table}[h!]
    \centering
    \resizebox{0.85\columnwidth}{!}{%
    \begin{tabular}{p{2cm}ll}
    \toprule
        \textbf{Resolution} & \textbf{Total Size of Videos (MB)} & \textbf{BodyMTS Accuracy} \\
        \midrule
        Default \newline (420 x 460)              & 213        & 0.87 \\
        One-half \newline (210 x 230)             & 48         & 0.82 \\
        One-third \newline (140 x 154)            & 27         & 0.75 \\
        \bottomrule
    \end{tabular}
    }
    \caption{Average accuracy of BodyMTS on test data over three train/test splits for different video frame  resolution.}
    \label{table:reso_accuracy}
    \end{table}
    
    \noindent\textbf{Reducing the bit-rate}: We alter the CRF  in order to modify the bit rate. CRF ranges from 0-53 and the default value of CRF is 23. We change the value of CRF with a step size of 6 as suggested in \citep{ffmpeg}, starting from 16. Resolution remains the same when changing the CRF. Table \ref{table:crf_accuracy} shows the impact of changing the CRF (or bit-rate) on the classifier accuracy.\\
    \begin{figure}
    \centering
    \includegraphics[width=0.9\columnwidth]{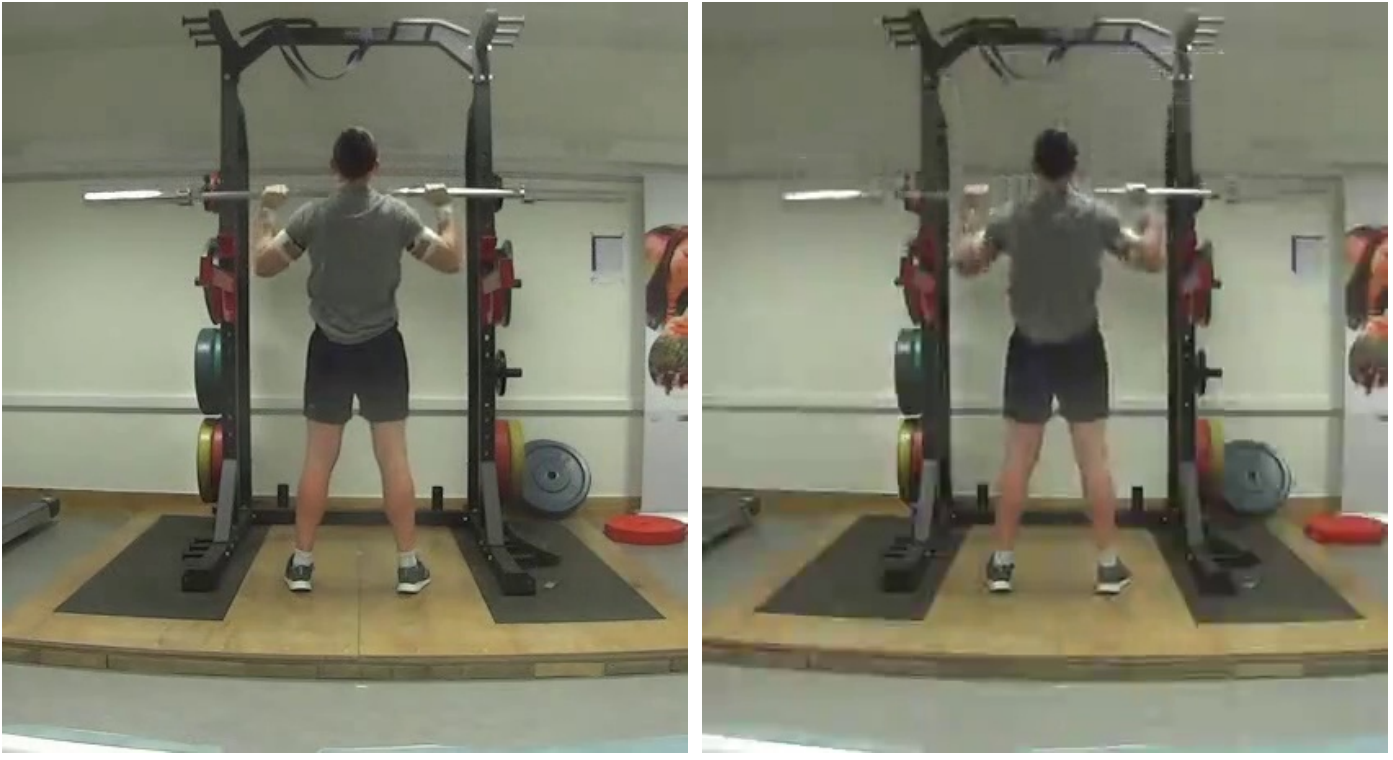}
    \caption{Single frame for class Normal at default resolution  of 420x460 and CRF 23 (default) versus CRF 40 (lower quality, distorted). The figure at CRF 40 (right) has a low resolution and hence looks blurrier than the figure at CRF 23 (left).}
    \label{fig:goodvsbad}
    \end{figure}
    \begin{table}[h!]
    \centering
    \resizebox{0.85\columnwidth}{!}{%
    \begin{tabular}{lll}
    \toprule
        \textbf{CRF}      & \textbf{Total Size of Videos (MB)} & \textbf{BodyMTS Accuracy} \\
        \midrule
        Default (23)      & 213  & 0.87 \\
        16                & 398  & 0.87  \\
        22                & 208  & 0.87  \\
        28                & 76   & 0.85 \\
        34                & 34   & 0.81 \\
        %40                & 20   & 0.68 \\
        % 46  & 0.51 & 13 \\
        \bottomrule
    \end{tabular}
    }
    \caption{Average accuracy of BodyMTS on test data over three train/test splits for different values of CRF. At CRF 28 we save 70\% of data storage and maintain similar  accuracy.}
    \label{table:crf_accuracy}
    \end{table}

    \noindent\textbf{Results and Discussion}: 
    We see in Table \ref{table:reso_accuracy} that reducing the resolution has a negative impact on the classifier accuracy. The reduction in average accuracy is more than 10 percentage points  when the original resolution is reduced to one-third. This confirms that degrading the quality of videos by reducing the resolution leads to a significant drop in accuracy. We are interested in finding good trade-offs between saving storage space and maintaining or only slightly sacrificing the accuracy. 
    % We also experimented with the color space by converting the default RGB to gray-scale and observed that BodyMTS accuracy remains unaffected.
    Next we alter the CRF property in order to change the bit-rate. Increasing the CRF leads to a drop in the bit-rate which leads to degraded quality of the videos and vice versa. Table \ref{table:crf_accuracy} shows that increasing the value of CRF leads to a drop in the classifier accuracy whereas decreasing this value has no effect on the accuracy. Figure \ref{fig:goodvsbad} shows a single frame for class Normal at CRF 23 (default) and CRF 40 where the image becomes too distorted to be usable.
    % This is directly correlated with the bit-rate where increasing the CRF leads to drop in the bit-rate which leads to degraded quality of the videos, hence the drop in the accuracy, whereas decreasing the CRF leads to an increased bit rate with no impact on the classifier accuracy. 
    The change in the video quality at CRF 22 and 23 is insignificant and hence the accuracy remains consistent. We observe that setting the value of CRF to 28 has no major impact on the classifier accuracy. This suggests that it is possible to reduce the total storage space of original videos while maintaining accuracy. The total size of the original videos is 213MB at CRF 23 but it is 76MB at CRF 28, hence a saving in storage space of 70\%. Additionally, the size of final time series is 28 MB which suggests further savings in storage space as compared to the original videos.
    
    \textbf{Takeaway:} Degrading the quality of videos by altering CRF to 28 makes it possible to satisfy minimum accuracy requirements (e.g. above 80\%) as listed in Table \ref{table:body_requirements} with 70\% savings in storage space. 
    % In conclusion, results from Table \ref{table:crf_accuracy} and Table \ref{table:reso_accuracy} shows that the quality of video plays a crucial role in determining the accuracy of BodyMTS.
    % The total execution time of BodyMTS in all the above different settings remains unaffected by changes in the video size and is around 2 minutes.
    
    % The classifier execution time in BodyMTS (data pre-processing + training + testing) for all the different parameters of video is around 2 mins and this remains unaffected by changes in the video size.\\

%%%%%%%%%%%%%%%%%%%%%%%%%%%%%%%%%%%%%%%%%%%
    
    \subsubsection{Results for Impact of Noise due to OpenPose Parameters}
    In this section, we analyze the impact of changing the OpenPose parameters as discussed in the previous section. Table \ref{table:pose_para_accuracy1} shows the accuracy for different parameter values and the total training and testing time of BodyMTS. Training time includes time taken for running OpenPose, data pre-processing and training the model, similarly testing time includes time taken for running OpenPose, data pre-processing and testing the model. Note that the impact of changing OpenPose parameters has been evaluated on the original dataset (default video settings for resolution and CRF). There are a total of 205 clips (2053 repetitions) of Military Press with a total combined duration of 1 hours 35 mins.\\

    \begin{table}[h!]
    \centering
    \resizebox{\columnwidth}{!}{%
    \begin{tabular}{p{6cm}lp{3cm}}
    \toprule
    \textbf{OpenPose parameters} & \textbf{BodyMTS Accuracy} &  \textbf{Training/Testing time (mins)} \\
        \midrule
        default \newline (frame-step=1,net-resolution\newline=-1x368,scale-number=1)        & 0.87  & 52/22 \\
        frame-step=2                  & 0.86  & 37/17 \\
        \textbf{frame-step=3}                  & \textbf{0.85}  & \textbf{26/12} \\\hline
        net-resolution=640            & 0.85  & 174/76 \\
        net-resolution=720            & 0.82  & 234/105 \\
        scale-number=4                & 0.82  & 140/63 \\
        \bottomrule
    \end{tabular}
    }
    \caption{Average accuracy on test data over three train/test splits for different OpenPose parameters. The total duration of all clips including training and testing clips is 95 mins.}
    \label{table:pose_para_accuracy1}
    \end{table}
    
    \noindent\textbf{Results and Discussion}: 
    From Table \ref{table:pose_para_accuracy1} we observe that increasing the \emph{frame-step} from 1 (using every frame) to 3 (every third frame), leads to a small drop in accuracy of 2 percentage points, but a significant reduction in run-time of OpenPose. 
    %This means that for pose estimation we do not need to consider every frame, and using every third frame is enough to capture the movement of body parts relevant for this classification task. 
    We further observe that increasing the values of the parameters \emph{net-resolution and scale}, which are mainly responsible for the confidence of OpenPose, produce no improvement on the accuracy, but rather leads to an increase in the overall run-time and a drop in the accuracy. These results suggest that the default values of these parameters are enough to give a reasonable accuracy using OpenPose. Table \ref{table:pose_para_accuracy1} also shows the total running time of OpenPose. Increasing the \emph{frame-step} makes OpenPose faster since it is not processing all the frames and so skipping some frames leads to faster execution time. The accuracy obtained by setting frame-step to 3 is still above the minimum desired accuracy of 80\% listed in Table \ref{table:body_requirements}, which means that for pose estimation we do not need to consider every frame, and using every third frame is enough to capture the movement of body parts relevant for this classification task. 
    
    \textbf{Takeaway:} By using a frame-step size of 3 along with default values of the remaining parameters for OpenPose, it is possible to noticeably reduce the training and testing time without a major drop in accuracy and still satisfy the minimum accuracy requirements in Table \ref{table:body_requirements}. 
    
    \begin{table}[h!]
    \centering
    \resizebox{0.95\columnwidth}{!}{%
    \begin{tabular}{p{6cm}lp{3cm}}
    \toprule
    \textbf{OpenPose parameters} & \textbf{BodyMTS Accuracy} \\
        \midrule
        default \newline (frame-step=1,net-resolution\newline=-1x368,scale-number=1) & 0.85 \\
        net-resolution=-1x640,\newline scale-number=4 scale-gap=0.25    &  0.85 \\
        net-resolution=-1x720,\newline scale-number=4 scale-gap=0.25    & 0.84 \\
        \bottomrule
    \end{tabular}
    }
    \caption{Average accuracy on poor quality test data at CRF 28 over three train/test splits for different OpenPose parameters.}
    \label{table:openpose_badquality}
    \end{table}
    
    We further tried tuning the \emph{net-resolution} on lower  quality video at CRF 28. Table \ref{table:openpose_badquality} shows the accuracy over different parameters of OpenPose for video quality at CRF 28. We observe no major improvement over the previous accuracy of 0.85. This also suggests that video quality plays a crucial role in determining the accuracy of BodyMTS as tuning OpenPose parameters alone is not sufficient to achieve good performance.
    % Moreover, in case of using every 2nd or 3rd frame, OpenPose can process faster than real time as the total combined duration of all the videos is 1 hour 35 minutes. 
    
    % The time series classifier runtime is negligible here; once the time series are extracted, it takes about 2 minutes to train and predict on the entire dataset. %However, since time series is re-sampled to same length there is no improvement in the execution time of ROCKET for different parameter values of OpenPose.
    
    \subsubsection{Training on Good Quality Videos and Testing on Poor Quality Videos}
    Previous results shown in Tables \ref{table:reso_accuracy},  \ref{table:crf_accuracy} and \ref{table:openpose_badquality}, considered the scenario when both the train and test video data are impacted by the same degree of level of noise. In this section, we consider another scenario when training is performed using the original high quality video whereas testing is performed using poor quality video. The poor quality videos are generated by altering the CRF value. Table \ref{table:crf_poor_accuracy} shows the accuracy of BodyMTS when trained on high quality videos and tested on poor quality videos. 
    
    \begin{table}[h!]
    \centering
    \resizebox{0.9\columnwidth}{!}{%
    \begin{tabular}{lll}
    \toprule
        \textbf{CRF}                   & \textbf{BodyMTS Accuracy} &  \textbf{Total Size of Videos (MB)}\\
        \midrule
        Original data (CRF=23)         & 0.87  & 213 \\
        28                             & 0.85  & 76 \\
        30                             & 0.83  & 56 \\
        34                             & 0.78  & 34 \\
        % 40                             & 0.53  \\
        \bottomrule
    \end{tabular}
    }
    \caption{Average accuracy of BodyMTS on test data for different values of CRF over three train/test splits. The training is performed on original data (CRF=23), while the testing is performed on poor quality video by altering the CRF from 23 to 34.}
    \label{table:crf_poor_accuracy}
    \end{table}

    \noindent\textbf{Results and Discussion}: We observe from Table \ref{table:crf_poor_accuracy} that BodyMTS accuracy drops when trained on high quality videos and tested on poor quality videos after CRF 30. From the above results it is clear that a threshold of CRF 30 can be chosen to reduce the data size while still satisfying the minimum accuracy requirements as listed in the Table \ref{table:body_requirements}. As previously observed in Table \ref{table:crf_accuracy} we save 70\% in storage space at CRF 28 without compromising on the accuracy.

    \subsubsection{Discussion on Impact of Video Quality and OpenPose Parameters on BodyMTS} In the above experiments, we studied the impact of two major sources of noise from video quality and the OpenPose parameters on the accuracy of BodyMTS. We observed that the quality of the videos has a large impact on the classifier accuracy. We also found that degrading the quality of videos by introducing a small amount of noise (at CRF 28) can lead to large savings for the storage space (from 213MB to 76MB), at a very small drop in accuracy (from 87\% to 85\%). This can be essential for applications deployed on low memory devices such as mobile phones, as well as where bandwidth is a constraint. From the pose estimation side we observed that the default values of OpenPose parameters are sufficient for good accuracy. The total duration of the original videos is 1 hour 38 minutes, whereas OpenPose took 1 hour 12 minutes using default parameters and this was further reduced to 38 minutes by utilizing every third frame during the estimation with a very small drop in accuracy (from 87\% to 85\%). Thus we can save both storage space and run-time, which is very promising given the constraints and requirements of this application. We note that there are many other types of video noise we can investigate, but we focused here on a subset of the most relevant sources for this application.
    %This suggests that OpenPose can process the data at a much faster frame rate than the original frame rate of 30 fps. %This makes OpenPose a most probable candidate for pose estimation.
    % Furthermore, we observe that the data pre-processing steps such as re-sampling and ROCKET's parameters have no major impact on the accuracy. We further presented the video quality metrics to quantify the amount of noise present in the videos.
   % We note that the noise types we studied is not a comprehensive list of sources of noise and so it may warrant further investigation depending upon the context of the application.

    \subsection{Robustness Analysis: Time Series Classification in BodyMTS}
    \label{sec:tscexp}
    
    In this section, we evaluate several multivariate time series classifiers and report the average accuracy over 3 train/test splits. 
    %We use 3 train/test splits for faster execution and to make it feasible to compare different settings for BodyMTS, as well as to compare BodyMTS with deep learning methods.
    We present results using both the previous version of OpenPose (v1.4) and the latest version (v1.7 at the time of writing). We are interested to see if the improvements in OpenPose lead to a significant improvement in the classification accuracy. For FCN and ResNet, we did not tune any hyperparameters and used the defaults as mentioned in the original papers \citep{Fawaz2019}. We observe that for ROCKET \citep{Dempster2020} changing the number of kernels (default=10000) did not produce any significant change on the accuracy, hence we kept the defaults as recommended by the authors. Where the algorithm allows it by exposing this option to the user, we disable the time series normalisation step. We show detailed results for varying data pre-processing in the Appendix.
    \begin{table}[h!]
    \centering
    \resizebox{0.95\columnwidth}{!}{%
    \begin{tabular}{lll}
    \toprule
        \textbf{Time Series Classifier} & \textbf{Accuracy OpenPose (v1.7)} & \textbf{Accuracy OpenPose (v1.4)} \\ 
        \midrule
        FCN & 0.82 $(\pm0.012)$ & 0.72 $(\pm0.043)$ \\
        ResNet & 0.76 $(\pm0.040)$ & 0.73 $(\pm0.028)$ \\
        ROCKET & \textbf{0.87} $(\pm0.026)$ & \textbf{0.81} $(\pm0.03)$ \\
        MINIROCKET & 0.81 $(\pm0.030)$ & 0.75 $(\pm1.644)$ \\
        \bottomrule
    \end{tabular}
    }
    \caption{Average accuracy on test data over 3 splits for multivariate time series classifiers trained with time series extracted with OpenPose version 1.4 and version 1.7.}
    \label{table:accuracy_stats}
    \end{table}
    
    \begin{table}[h!]
    \centering
    \resizebox{0.9\columnwidth}{!}{%
    \begin{tabular}{lll}
    \toprule
        \textbf{Time Series Classifier} & \textbf{Training Time (mins)} & \textbf{Testing Time (mins)} \\ 
        \midrule
        FCN & 85 & 28  \\
        ResNet & 115  & 28 \\
        ROCKET &  \textbf{52} &  \textbf{22} \\
        MINIROCKET & 50 & 20 \\
        \bottomrule
    \end{tabular}
    }
    \caption{Average time taken for training and testing over 3 splits for the selected methods. Note that training and testing time shown here is inclusive of the time taken for OpenPose and data pre-processing. \textbf{The total duration of all clips in training and testing is around 65 mins and 30 mins respectively.}}
    \label{table:time_classifiers}
    \end{table}

    We also present results for different subsets of time series dimensions (i.e., body parts). We utilize the ECP method \citep{dhariyalfast} for automated dimension selection due to the large number of possible combinations from the 25 body parts (50 dimensions). We use both the left and right parts for a single body part unless otherwise stated. We further compare this with the number of dimensions suggested by the domain experts who carried out the data collection. 
    % Based on the above results, we choose ROCKET as the main classifier in BodyMTS as it the most accurate and scalable method among the selected classifiers.
    
    \begin{table}[h!]
    \centering
    \resizebox{\columnwidth}{!}{%
    \begin{tabular}{lp{4cm}l}
    \toprule
        \textbf{Number of body parts} &  \textbf{Body parts}                 & \textbf{ROCKET Accuracy}  \\
        \midrule
        25 body parts                 & all                       & 0.83 $(\pm0.028)$  \\
        8 upper body parts (domain expert)            & wrists, elbows, shoulders and hips    & \textbf{0.87} $(\pm0.026)$  \\
        8 body parts (\cite{dhariyalfast})                  & wrists, elbows, big toes and small toes  & 0.81 $(\pm0.012)$  \\
        6 upper body parts                  & wrists, elbows and shoulders         &  \textbf{0.87} $(\pm0.021)$  \\
        4 upper body parts                  & wrists and elbows                    & 0.85 $(\pm0.012)$  \\
        \bottomrule
    \end{tabular}
    }
    \caption{Average accuracy of ROCKET using OpenPose v1.7 on test data for different subsets of body parts over 3 train/test splits.}
    \label{table:accuracy_stats_diff_body_parts}
    \end{table}
    
    \noindent\textbf{Results and Discussion:}
    Table \ref{table:accuracy_stats} shows the average accuracy and standard deviation obtained on the test data over three data splits. ROCKET achieved the highest accuracy for both versions of OpenPose, followed by the deep learning models. The standard deviation values are generally small, which means accuracy generally remains consistent over different splits. We further observe that there is a notable increase of 6 percentage points in the accuracy of ROCKET transitioning from the older version to the latest version of OpenPose (v1.7 at the time of writing this paper). We think that further improvements in  OpenPose and time series classifiers can further improve the performance on this task. With an average accuracy of 87\%, it confirms that OpenPose coupled with ROCKET can surpass the minimum accuracy requirement of 80\% as listed in  Table \ref{table:body_requirements}. 
    
    Table \ref{table:time_classifiers} shows the training time and testing time for the selected methods. We observe that MINIROCKET takes the least amount of time for training and testing followed by ROCKET and deep learning methods. However, despite MINIROCKET being faster than ROCKET in total time taken, it is less accurate than ROCKET as shown in Table \ref{table:accuracy_stats}.
    
    Based on the above results, we choose ROCKET as the classifier in BodyMTS as it the most accurate and scalable method among the classifiers investigated.
    In Table \ref{table:accuracy_stats_diff_body_parts} we observe the results using different subsets of body parts selected from the multivariate time series. We note that careful selection of the body parts has a significant impact on the accuracy of the classifier, with the upper body parts recommended by the domain expert achieving the highest accuracy among alternative subsets considered.  The automated selection method proposed by \citep{dhariyalfast} also selects 8 body parts, but seems to be affected by data noise in selecting lower information body parts such as toes, which do not play a role in accurately classifying the movement. 

\section{Lessons Learned and Limitations of the BodyMTS Approach}
\label{sec:discussion}

    BodyMTS  achieves an accuracy of about 87\% on the Military Press video dataset. Based on these results, it is possible to use videos as an alternative to sensor-based approaches for human exercise classification. However, further work is needed to analyze the generalizability of BodyMTS by including other strength and conditioning exercises. We analyzed the robustness of BodyMTS against common sources of noise, particularly the video quality and OpenPose parameters. We observed that video quality plays a critical role in determining the classification accuracy. We showed that video quality can be degraded to a CRF value of 28 without a significant drop in accuracy, whilst achieving savings in terms of storage space. We observed that the subset of channels (i.e., body parts) has a large impact on classier accuracy, which necessitates further investigation. We have also seen that improvements in the body pose estimation method (from OpenPose v1.4 to v1.7) have resulted in higher accuracy, which is encouraging. Furthermore, in case of data pre-processing, segmentation plays a crucial role, as incorrect peak detection due to a noisy signal may lead to incorrect capturing of a repetition, which in turn affects the classifier accuracy. Lastly, due to the choice of ROCKET as a classifier, BodyMTS does not involve tuning too many parameters. Hence BodyMTS as an application requires setting fewer parameters. Finally, we compared BodyMTS with state-of-the-art deep learning methods and observed that BodyMTS achieves better accuracy while requiring fewer computation resources. Additionally, a lot of engineering effort is required to tune the deep learning models.
    %, whereas BodyMTS requires no more than 3 parameters (\emph{frame-step} for OpenPose and \emph{normalization}).
    
    Below we discuss some of the limitations and mitigations for this approach.
    
    \begin{itemize}
    \item \textbf{Types of Exercises and Pose Estimation.} BodyMTS may fail in case of exercises where the participants may have to lie down, e.g., for exercises like push-ups and sit-ups. In such cases, videos may fail to capture the full motion or the pose estimation may fail due to body parts occlusion. Further, we observed in the previous experiments \citep{ashish2020} that the classifier struggles to classify some samples from class Normal and Arch for  Military Press, which is due to the fact that the front view may not be able to fully capture the lateral motion. 
    % However, usage of multiple cameras to record the execution can overcome these issues \citep{nakano2020evaluation}. 
    Techniques like extending the 2D information to 3D through triangulation techniques \citep{Kwon2020, nakano2020evaluation} can be used to capture the depth information. BodyMTS is also limited by the number of body parts detected by OpenPose. Physical exercises which require tracking of certain body parts not supported by OpenPose may not be well suited for this approach. For instance, for an exercise which requires to track all vertebrae of the spine, it may not be possible with this approach as OpenPose does not currently track each vertebra of the spine. However, most types of Strength and Conditioning exercises involve majorly the upper or lower body parts movements which are tracked well with OpenPose.
    
    \item \textbf{Video Data Capture.} 
    Factors such as lighting, viewing angle, stability and position of the camera, camera quality, clothing of the participants, background, location, etc., can possibly influence the video recordings of the exercise. OpenPose may fail to detect some body parts if the body part is not fully captured in the video recordings. Additionally, the confidence of OpenPose is directly related to the video quality. The lower quality of the videos leads to decrease in the accuracy of OpenPose and hence low quality of final data. However, issues like video quality can be easily overcome by usage of smartphones that have a reasonable quality camera, whereas accidental cropping of body parts as well as positioning and the viewing angle can be avoided by requirements for the video recordings as mentioned in  Section \ref{sec:app_require}. OpenPose is robust against occlusions including during human-object interaction \cite{openpose2019}. 
    %and can easily handle real-time pose estimation for multiple persons. 
    Moreover, lightweight versions of OpenPose \citep{osokin2018lightweight_openpose} and TensorFlow Lite\footnote{\url{https://www.tensorflow.org/lite/examples/pose_estimation/overview\#performance_benchmarks}} makes it possible to run pose estimation frameworks on resource-constrained devices such as smartphones as well as on a single CPU without sacrificing the accuracy. Lastly, recent works based on preserving the privacy and security of participants \citep{hinojosa2021learning} do not require participants to be completely visible in front of the cameras thus preserving their privacy which is in contrast to the deep learning methods that might require that participants be completely visible.
    
    % Lastly, OpenPose has already been adopted as the pose estimation library across many organizations.

    \item \textbf{Data Pre-processing.} BodyMTS uses peak detection to segment the individual time series from multiple repetitions to obtain single repetitions. Due to noise, jittering and depending on the accuracy of OpenPose, segmentation may not always be correct and hence may lead to loss of some repetitions. Additionally, the duration of a single repetition cannot be generalized as it varies from participant to participant. In our experiments only a few samples were dropped due to incorrect segmentation without having a substantial impact on the accuracy. 
    % Also, as we saw earlier in the Table \ref{table:accuracy_stats_diff_body_parts} that including all the body parts lead to a drop in the accuracy of BodyMTS. This warrants further investigation to select the correct number of dimensions as it may not be always possible to consult the domain experts.
    % Moreover, BodyMTS due to its lightweight nature has a very low number of hyper parameters to tune, thus saving a lot of engineering effort as opposed to deep learning based methods. 

    % Possible solutions to overcome this is to manually analyze the data and mark the length of each rep which is time-consuming. 
    
    \end{itemize}

    Despite these limitations, our experiments show that BodyMTS is a very promising approach. 
    %It is a fast, robust and accurate method to classify the performance of video-based S\&C exercises. 
    Advancements in body pose estimation and time series classification may help in further improving the performance of BodyMTS. An interesting future direction concerns the development of time series classification algorithms that are more flexible and do not require strict pre-processing steps such as length re-sampling or normalisation. 
    Additionally, the data produced by OpenPose has associated pose estimation confidence values, and this raises interesting research questions of how the classifier may benefit from knowledge of uncertainty in the data to improve the accuracy.
    %Model & Openpose time & Classifiation time & Total &hit@1 \\ \midrule

\section{Recommendations for Practitioners}
\label{sec:recomm_prac}

We divide this section into three parts: video data, pose estimation and time series classification. We provide recommendations for each of these based on the experiments performed and our findings.

 \begin{itemize}
    \item \textbf{Using videos as data source for S\&C exercise classification.}
    %The camera view used to record the exercises varies depending upon the exercise. BodyMTS utilizes the front view of the camera for Military Press and achieves an accuracy of 87\%. Also, camera from the readily available smartphones can be utilized to record the execution.
    We recommend the use of videos as an alternative source to sensors, for S\&C exercise classification. The large storage and computation requirements of video can be addressed by changing the bit rate property, which reduces the video quality and size. In the case of Military Press, a CRF up to about 28 to 30, reduces the data size by 70\% without affecting the accuracy.

    Certain application requirements on the video quality and data capture process need to be met. For example, the requirement of a stable camera, as well as the participant being fully captured in each frame are important to ensure the quality of the follow up steps.
    We have detailed such requirements in Section \ref{sec:app_require}.
    
 %The distance between the participants and the cameras does not remain constant for different participants. The camera should be placed on a static, horizontal surface and the participant should be completely captured by the camera. 
    We recommend that the recorded videos are then  pre-processed to remove any audio, background, and centralize the participants using a bounding box, followed by changing the video CRF value which results in saving storage space.
    
    \item \textbf{Pose Estimation.} We recommend the use of  OpenPose for human pose estimation. It can process video faster than real-time, and has parameters such as \emph{frame-rate} which allows skipping frames during processing. We have found that skipping 1-2 frames further improves speed and does not affect the accuracy of the classification step. The body parts detection and tracking is sufficiently accurate to enable the followup classification step. We found that improvements in pose estimation in v1.7 have also lead to improvements in classification.
    %as it supports real time multi person pose detection along with many configurable parameters. 
    %In our experiments, we utilize \emph{number-people-max} to limit the number of participants and \emph{frame\-rate} to quickly process the videos. 
    %Tuning parameters like \emph{net-resolution, scale-number} did not result in any improvement and hence should be set to their default values along with remaining parameters. It should also be noted that O
    %OpenPose confidence is directly proportional to the video quality and therefore any degradation of the video quality will result decrease in the OpenPose' confidence and ultimately a drop in the classification accuracy. 
    Libraries such as OpenVINO \footnote{\url{https://docs.openvino.ai/2019_R1/_human_pose_estimation_0001_description_human_pose_estimation_0001.html}} make it possible to execute OpenPose on  CPU, and TensorFlow Lite \footnote{\url{https://www.tensorflow.org/lite/examples/pose_estimation/overview\#performance_benchmarks}} supports running pose estimation models directly on a mobile phone in real-time, so there is a lot of promise in current developments for pose estimation.
    
    \item \textbf{Data pre-processing and multivariate time series classifiers.}
    We recommend using pose estimation information to obtain the human motion time series and then segment the multiple repetitions of the exercise. This is more accurate than other heuristics for segmenting the reps.
    %The data obtained after pose estimation is segmented to obtain time series for each repetition. 
    Before segmentation, it helps if the full signal is smoothed using a Savgol filter followed by peak detection. %Indices of peaks are recorded and the time points from one peak to next consecutive peak in the raw signal are used to create single rep records for train/test data. 
    We recommend to not normalise the data of each repetition, as this can change the meaning of the data and also decreases the  classification accuracy.
    %We observe that setting normalization to True leads to a significant drop in the accuracy. Finally. each time series is resampled to same length as the BodyMTS does not support variable length time series data. Moreover, 
    We found that length re-sampling does not impact the final classification accuracy.
    We recommend ROCKET for the multivariate time series classification step, as it is very fast and most accurate among the classifiers evaluated. 
    %The normalization flag in the Ridge Classifier is set to True for faster convergence. All the parameters can be set to default as there is no major impact on the accuracy.
    
 \end{itemize}
\section{Conclusion}
\label{sec:conclusion}
In this work we have analyzed the performance and robustness of BodyMTS, an approach for exercise classification using videos as time series.% by transforming the videos into body pose tracking time series. 
We presented the required features and associated technical requirements for this kind of application. We evaluated BodyMTS on a real-world dataset for the Military Press exercise and achieved an average accuracy of 87\%. We further showed that the latest improvements in body pose estimation with OpenPose can improve the performance of BodyMTS. We observed that the subset of channels (i.e., body parts) has a large impact on classifier accuracy, which necessitates further investigation in future work. We compared the robustness of BodyMTS against different sources of noise particularly related to variations in video quality and OpenPose parameters. We observed that BodyMTS can achieve good performance at different levels of video quality. 
%The results showed that the quality of videos has a large impact on the accuracy of OpenPose, which ultimately affects the final classifier accuracy. 
We showed that by decreasing the quality of videos, a major portion of storage cost can be reduced (70\%), with a very small drop in accuracy (2 percentage points). This leads to less computation as well as savings in terms of storage cost. We further noticed that changing the parameters of OpenPose has less impact on the classifier accuracy, but can lead to large savings in running time. Lastly, we compare the BodyMTS approach with deep learning-based methods for human activity recognition from videos. We considered several aspects such as performance, storage space and practicality. We observed that BodyMTS can achieve better performance than deep learning methods in terms of total time and accuracy, without the use of heavy computing resources. For future work, we plan to carry out extensive experiments to analyze how BodyMTS generalises to more strength and conditioning exercises. Additionally, we plan to compare the performance of BodyMTS with sensors-only based approaches. 
% The sensor data is already available for the same participants who participated in the Military Press study. 
Furthermore, we plan to work on the classification interpretation aspect, where the goal is to provide useful prediction explanation feedback to the end-user.

% Extend the conclusion GI

\section{Appendix}
\label{sec:appendices}

    \subsection{Time Series Data Pre-processing and Classification}
    We consider the impact of time series length re-sampling and data normalization here. We use a single split of data here for faster execution.
    
    \begin{itemize}
    \item \textbf{Re-sampling}: Each time series  has been re-sampled to the same length since most time series classifiers cannot handle variable-length data. 
    
    \item \textbf{Normalization}: The magnitude of the signal is important for this application, as shown in the experiment here.
    
    \end{itemize}
    
    \begin{table}[h!]
    \centering
    \resizebox{0.7\columnwidth}{!}{%
    \begin{tabular}{ll}
    \toprule
    \textbf{Re-sampling Length} & \textbf{BodyMTS Accuracy } \\
        \midrule
%        25   & 0.83 \\
        50   & 0.81 \\
        100  & 0.83 \\
        161  & 0.83 \\
        225  & 0.83 \\
        300  & 0.83 \\
        500  & 0.83 \\
    \bottomrule
    \end{tabular}
    }
    \caption{Accuracy on test data over a single train/test split for different values of time series length.}
    \label{table:resam_accuracy}
    \end{table}
    
    \begin{table}[h!]
    \centering
    \resizebox{0.7\columnwidth}{!}{%
    \begin{tabular}{ll}
    \toprule
    \textbf{Normalized data} & \textbf{BodyMTS Accuracy } \\
        \midrule
        True        & 0.84 \\
        False       & 0.88 \\
    \bottomrule
    \end{tabular}
    }
    \caption{Impact of normalization on accuracy using test data over 3 train/test split.}
    \label{table:norm_unoorm_accuracy}
    \end{table}
    
    \noindent\textbf{Results and Discussion}: Table \ref{table:resam_accuracy} shows the impact of changing the re-sampling length on the BodyMTS accuracy. We see that there is almost no effect of length re-sampling on the classifier accuracy. Furthermore, reducing the length of the data also leads to reduced execution time of BodyMTS. We also experimented with the parameters of the ROCKET classifier such as number of kernels (10000) and normalization (False). While changing the number of kernels did not produce any impact on the accuracy, setting the normalization to FALSE lead to a big increase in the accuracy as shown in Table \ref{table:norm_unoorm_accuracy}. We believe that this is due to the loss of magnitude information which is a key element in the classification for this type of problem. We further experimented by converting the color scale of videos to gray and observed no change in the accuracy of BodyMTS.
    
     \subsection{Quantifying Video Quality Noise using Video Quality Metrics} We further quantify the impact of noise on the classifier accuracy by using video quality metrics. We use three scores: VMAF \citep{netflixvmaf}, PSNR, and SSIM. FFmpeg has been utilized to calculate these metrics for different CRF values. We report the average metric score over all the clips for each value of CRF. Table \ref{table:metricsvideo_accuracy} shows the average score over all the clips and the accuracy obtained using a particular value of CRF.
      \begin{table}[h!]
      \centering
      \resizebox{\columnwidth}{!}{%
      \begin{tabular}{lllll}
      \toprule
      \textbf{CRF} & \textbf{Average VMAF score} & \textbf{Average PSNR} &  \textbf{Average SSIM} & \textbf{BodyMTS Accuracy } \\
          \midrule
          16 & 97.03 & 48.14 & 0.998 &  0.83 \\
          22 & 95.54 & 44.51 & 0.996 &  0.83 \\
          28 & 91.21 & 39.85 & 0.990 &  0.82 \\
          \midrule
      34 & 87.76 & 36.52 & 0.979 &  0.75 \\
          40 & 67.94 & 33.09 & 0.954 &  0.69 \\
          %46 & 43.99 & 29.63 & 0.900 &  0.51 \\
          \bottomrule
      \end{tabular}
      }
      \caption{Accuracy and video quality metrics score on test data over a single train/test split for different values of CRF.}
     \label{table:metricsvideo_accuracy}
      \end{table}

     \noindent\textbf{Results and Discussion}: We observe that the VMAF score is more useful than other scores for estimating the quality of videos. Higher VMAF, indicates a better quality of videos. There is a big drop in the average VMAF score by changing the CRF values. Based on these results the threshold of VMAF can be set at around 90 which can be used to exclude those videos whose VMAF score is less than 90.
     
    % \subsection {Two stage classification}

    %\subsection{Visualization of OpenPose confidence}
    %Here we show the average confidence for each exercise type for each participant for the right elbow. These confidence scores have been averaged over all the repetitions and all the classes for each participant. We only show few participants because of the space constraint. Figure \ref{fig:avg_confidence_openpose} shows this graph for participants belonging to testing data. 

    %\begin{figure}[ht]
    %    \centering
    %    \includegraphics[width=0.7\linewidth]{figs/rq2/average_confidence.pdf}
        %\caption{Average confidence for right elbow for each participant over all repetitions and classes.}
     %   \label{fig:avg_confidence_openpose}
    %\end{figure}
    
    %We see that decreasing the quality of video whether by increasing the CRF or reducing the resolution has a huge impact on the confidence of OpenPose which ultimately leads to a drop in the classifier accuracy.

\begin{acknowledgements}
This work was funded by Science Foundation Ireland through the Insight Centre for Data Analytics (12/RC/2289\_P2) and VistaMilk SFI Research Centre
(SFI/16/RC/3835).
\end{acknowledgements}

% \bibliographystyle{spbasic}      % mathematics and physical sciences
% \bibliography{main}   % name your BibTeX data base

\end{document}